\documentclass[twocolumn]{bmcart}
\usepackage{amsmath,amssymb,amsfonts}
\usepackage{algorithmic}
\usepackage{graphicx}
\usepackage{textcomp}
\usepackage{threeparttable}
\usepackage{multirow}
\MakeRobust{\overrightarrow}
\usepackage{times}
\usepackage{latexsym}
\usepackage{xurl}
\usepackage[linesnumbered,ruled,vlined]{algorithm2e}

\startlocaldefs
\definecolor{darkgreen}{rgb}{0.44,0.68,0.28}
\definecolor{darkblue}{rgb}{0,0.44,0.75}
\definecolor{mediumgrey}{rgb}{0.65,0.65,0.65}
\SetKwInput{KwRequire}{Require}
\SetKwInput{KwEnsure}{Ensure}
\endlocaldefs

\begin{document}
\begin{frontmatter}
	
\begin{fmbox}
\dochead{Research}

\title{Ontology-Driven and Weakly Supervised Rare Disease Identification from Clinical Notes}

\author[addressref={cmi,hdr,ox},corref={cmi,hdr,ox},email={hang.dong@cs.ox.ac.uk}]{\fnm{Hang} \snm{Dong}}
\author[addressref={cmi,hdr}]{\fnm{V\'{i}ctor} \snm{Su\'{a}rez-Paniagua}}
\author[addressref={acrc}]{\fnm{Huayu} \snm{Zhang}}
\author[addressref={ucl}]{\fnm{Minhong} \snm{Wang}}
\author[addressref={acrc}]{\fnm{Arlene} \snm{Casey}}
\author[addressref={ccbs}]{\fnm{Emma} \snm{Davidson}}
\author[addressref={man}]{\fnm{Jiaoyan} \snm{Chen}}
\author[addressref={efi}]{\fnm{Beatrice} \snm{Alex}}
\author[addressref={hdr,ccbs}]{\fnm{William} \snm{Whiteley}}
\author[addressref={hdr,ucl},corref={ucl},email={honghan.wu@ucl.ac.uk}]{\fnm{Honghan} \snm{Wu}}


\address[id=cmi]{\orgdiv{Centre for Medical Informatics, Usher Institute of Population Health Sciences and Informatics}, \orgname{University of Edinburgh}, \city{Edinburgh}, \cny{United Kingdom}}
\address[id=hdr]{\orgname{Health Data Research UK}, \city{London}, \cny{United Kingdom}}
\address[id=ox]{\orgdiv{Department of Computer Science}, \orgname{University of Oxford}, \city{Oxford}, \cny{United Kingdom}}
\address[id=acrc]{\orgdiv{Advanced Care Research Centre, Usher Institute}, \orgname{University of Edinburgh}, \city{Edinburgh}, \cny{United Kingdom}}
\address[id=ucl]{\orgdiv{Institute of Health Informatics}, \orgname{University College London}, \city{London}, \cny{United Kingdom}}
\address[id=ccbs]{\orgdiv{Centre for Clinical Brain Sciences}, \orgname{University of Edinburgh}, \city{Edinburgh}, \cny{United Kingdom}}
\address[id=man]{\orgdiv{Department of Computer Science}, \orgname{The University of Manchester}, \city{Manchester}, \cny{United Kingdom}}
\address[id=efi]{\orgdiv{Edinburgh Futures Institute}, \orgname{University of Edinburgh}, \city{Edinburgh}, \cny{United Kingdom}}


\begin{abstractbox}
\begin{abstract}
\parttitle{Background}
Computational text phenotyping is the practice of identifying patients with certain disorders and traits from clinical notes. Rare diseases are challenging to be identified due to few cases available for machine learning and the need for data annotation from domain experts. 
\parttitle{Methods}
We propose a method using ontologies and weak supervision, with recent pre-trained contextual representations from Bi-directional Transformers (e.g. BERT). The ontology-driven framework includes two steps: (i) Text-to-UMLS, extracting phenotypes by contextually linking mentions to concepts in Unified Medical Language System (UMLS), with a Named Entity Recognition and Linking (NER+L) tool, SemEHR, and weak supervision with customised rules and contextual mention representation; (ii) UMLS-to-ORDO, matching UMLS concepts to rare diseases in Orphanet Rare Disease Ontology (ORDO). The weakly supervised approach is proposed to learn a phenotype confirmation model to improve Text-to-UMLS linking, without annotated data from domain experts. We evaluated the approach on three clinical datasets, MIMIC-III discharge summaries, MIMIC-III radiology reports, and NHS Tayside brain imaging reports from two institutions in the US and the UK, with annotations.
\parttitle{Results}
The improvements in the precision were pronounced (by over 30\% to 50\% absolute score for Text-to-UMLS linking), with almost no loss of recall compared to the existing NER+L tool, SemEHR. Results on radiology reports from MIMIC-III and NHS Tayside were consistent with the discharge summaries. The overall pipeline processing clinical notes can extract rare disease cases, mostly uncaptured in structured data (manually assigned ICD codes).
\parttitle{Conclusion}
The study provides empirical evidence for the task by applying a weakly supervised NLP pipeline on clinical notes. The proposed weak supervised deep learning approach requires no human annotation except for validation and testing, by leveraging ontologies, NER+L tools, and contextual representations. The study also demonstrates that Natural Language Processing (NLP) can complement traditional ICD-based approaches to better estimate rare diseases in clinical notes. We discuss the usefulness and limitations of the weak supervision approach and propose directions for future studies.
\end{abstract}


\end{abstractbox}
\end{fmbox}
\end{frontmatter}

\section*{Introduction}
\label{intro}
Text phenotyping is the task of extracting diseases or traits of patients from clinical notes, which can benefit a wide range of tasks like cohort selection, epidemiological research, and decision making for better clinical care. A particular set of human phenotypes are rare diseases: a rare disease is very uncommon, affecting 5 or fewer people in 10,000, but there are between 6,000 and 8,000 rare diseases and they collectively affect approximately 3.5-5.9\% of the population (or 263–446 million persons)
globally \cite{nguengang2020estimating} (and over 1 in 17 people in the UK \cite{uk_gov_2021} and 8\% of population in Scotland \cite{scot_gov_2021}) at some point in their lifetime. Compared to common diseases, rare diseases are usually not coded in a precise manner, this is partly because they are under-represented in the current, ICD-10 (International Classification of Diseases, version 10) terminologies \cite{richesson2014,Bearryman2016}. Detailed information about a patient is usually hidden in unstructured, clinical narratives. It is thus necessary to use clinical notes with Natural Language Processing (NLP) techniques to complement coded data to identify rare diseases in patients.

The main challenge for rare disease identification with NLP is the lack of annotated data for machine learning, especially deep learning. Deep learning models for clinical note classification tend to perform worse for infrequent diseases due to the lack of cases for training \cite{dong2021}. On the other hand, annotating a variety of rare diseases in clinical notes from scratch needs specific domain expertise. This also requires the manual annotation of a very large number of clinical notes to ensure enough cases for each rare disease, thus taking time and incurring considerable costs from a group of clinical experts.

\begin{figure*}
  \centering
  \includegraphics[width=0.95\textwidth]{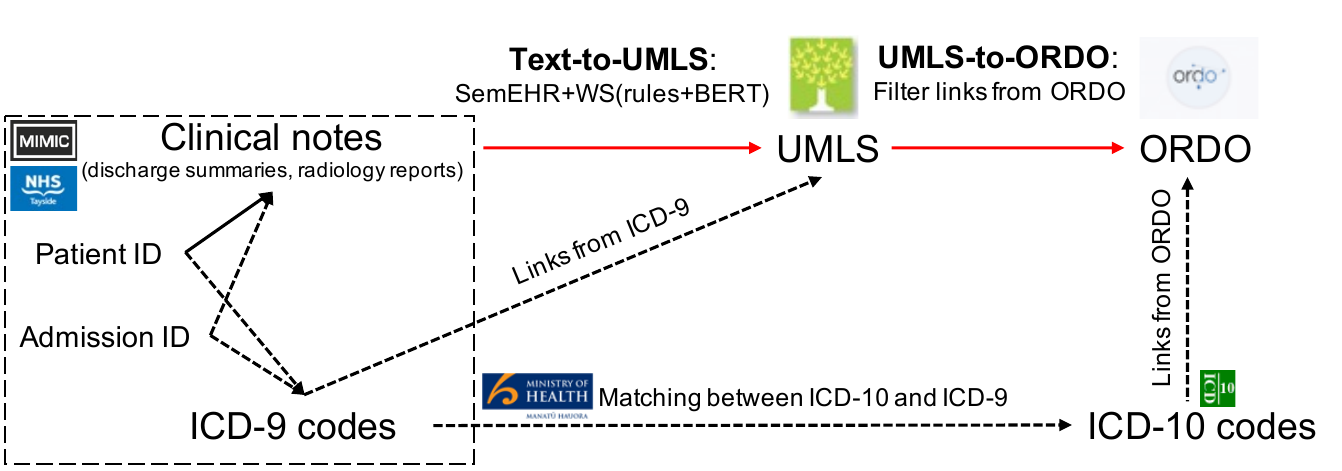}
  \caption{A pipeline for rare disease identification from clinical notes with ontologies and weak supervision. The upper horizontal lines (in \textcolor{red}{red}) show the proposed pipeline based on clinical notes (e.g. discharge summaries and radiology reports in US MIMIC-III and UK NHS Tayside) and ontologies, including two steps (Text-to-UMLS and UMLS-to-ORDO). No annotation data are needed, through a UMLS extraction tool, SemEHR, and weak supervision (WS) based on customised rules and BERT-based contextual representations (see details on WS in Figure \ref{pipeline-weak}). The admission ID and ICD-9 codes (linked with dotted lines) are only available for the MIMIC-III data. The lower, dotted lines show a baseline approach purely based on manual ICD codes, also enhanced with ontology matching. (Figure adapted from \cite{dong2021rare}.)}\label{pipeline-main}
\end{figure*}

We propose an ontology-driven and weakly supervised framework for rare disease identification from clinical notes, extending our previous work in \cite{dong2021rare} with further, detailed empirical analyses and external validation. Ontologies are essential for text phenotyping as they provide a curated list of terms of diseases and traits. Previous studies have used ontologies to estimate the frequency of rare diseases \cite{kahn2017ontology}. Our main ontology-driven framework is illustrated in Figure \ref{pipeline-main}. 

We use Orphanet Rare Disease Ontology \cite{vasant2014ordo} as the list of vocabularies of rare diseases\footnote{We focus on the identification of diseases instead of the associated phenotypic abnormalities, therefore we chose ORDO instead of Human Phenotype Ontology (HPO) \cite{GROZA2015hpo}. The overall ontology based and weak supervision framework can potentially be applied to HPO, given it being aligned to ORDO \cite{MAIELLA2018}. We leave the phenotypic abnormalities (in HPO) for future studies.}. We then leverage the concepts and synonyms in Unified Medical Language System (UMLS) as an intermediary dictionary to extend matching terms and address the issue of name variation \cite{shen2015} in linking texts to rare diseases, e.g. ``tracheobronchomalacia'' for Williams-Campbell syndrome. The framework thus contains two integrated parts, \textit{entity linking} (Text-to-UMLS) and \textit{ontology matching} (UMLS-to-ORDO). Entity linking from mentions (or text fragments) to UMLS concepts is challenging due to the ambiguous mentions \cite{shen2015,kahn2017ontology}, especially for abbreviations, e.g. ``HD'' which could mean Huntington Disease, Hemodialysis, or Hospital Day. String matching usually does not consider the complex contexts of a mention and can therefore result in many false positives. Machine learning can be applied for the disambiguation of terms, but it needs abundant annotated training data, which are currently not available in the context of rare diseases.

We therefore propose a weakly supervised approach to filter out the false positives in entity linking. Weak supervision \cite{wang_clinical_2019,ratner2019} is a strategy to automatically create labelled training data using heuristics, knowledge bases, crowdsourcing, and other sources, to alleviate the burden and cost of annotation. We first use a string matching based named entity linking tool, SemEHR \cite{Wu2018semehr} (widely applied for text phenotyping in the UK \cite{Wu2018semehr,wu2019,gorinski_named_2019}, based on Bio-YODIE \cite{gorrell2018}) to generate candidate entity linking results, i.e. mentions and their UMLS concepts, from clinical notes; then, we propose to efficiently create weak training data of candidate mention-UMLS pairs of sufficient quality with two rules, mention character length, regarding ambiguous abbreviations, and ``prevalence'', regarding rare diseases. A phenotype confirmation model can thus be learned through contextual mention representations with domain-specific BERT models (e.g. BlueBERT \cite{peng2019transfer}) to capture the context under-lied in the texts to disambiguate the mention to improve entity linking. For UMLS-to-ORDO matching, we used the mappings in ORDO and corrected the wrong links by filtering ORDO concepts with a phenome type as an upper class in the ontology \cite{vasant2014ordo}. 

\begin{figure*}[ht]
  \centering
  \includegraphics[width=0.8\textwidth]{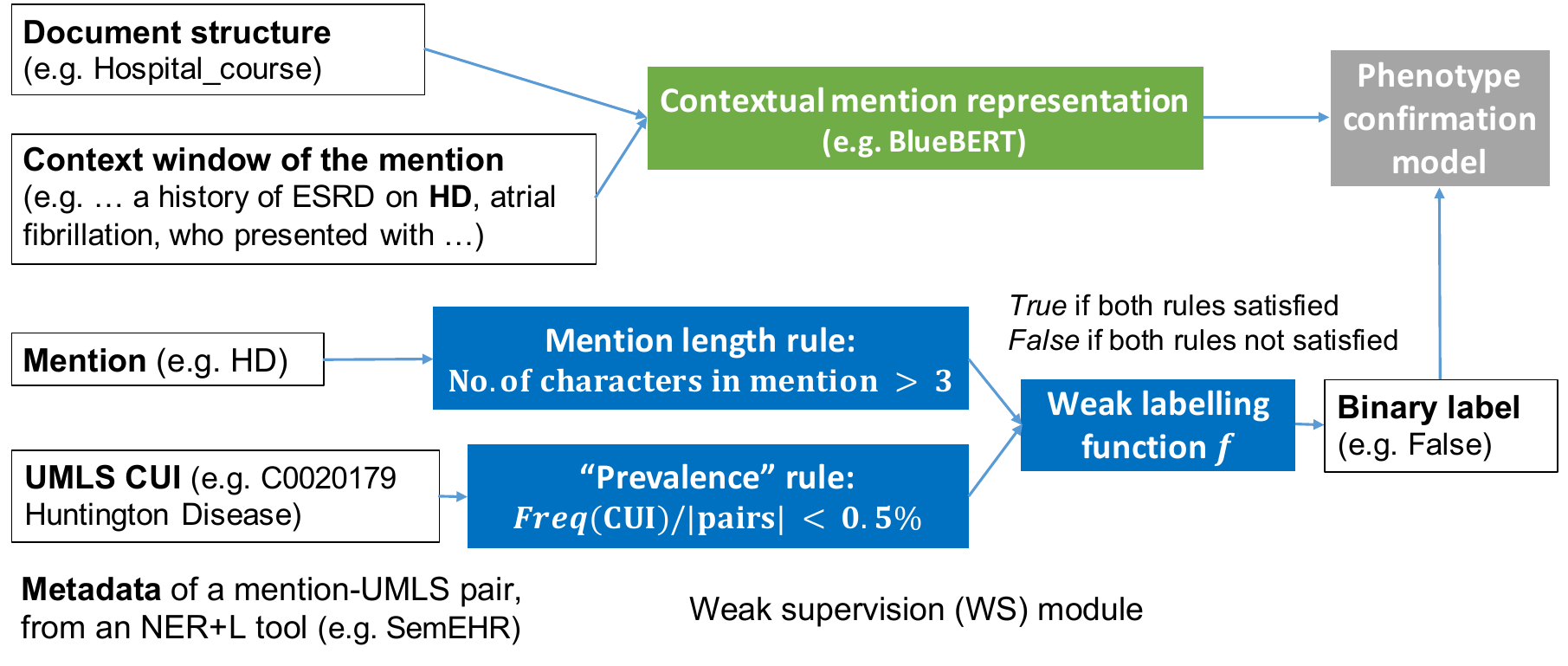}
  \caption{Weak supervision process for Text-to-UMLS linking. The left four white text boxes displayed the metadata (with examples) of a candidate mention-UMLS pair, identified by a Named Entity Recognition and Linking (NER+L) tool, SemEHR; the coloured text boxes in the middle show the contextual representation block (in \colorbox{darkgreen}{\textcolor{white}{\textbf{green}}}) and the rule-based weak data labelling (in \colorbox{darkblue}{\textcolor{white}{\textbf{blue}}}). A binary label is then generated, which weakly estimates whether the candidate pair indicates a correct phenotype of the patient. A phenotype confirmation model (in \colorbox{mediumgrey}{\textcolor{white}{\textbf{grey}}}) is then learned to select correct phenotypes from the pairs. (Figure adapted from \cite{dong2021rare}).)}\label{pipeline-weak}
\end{figure*}

For our main experiments, we trained a weakly supervised phenotype confirmation model using the discharge summaries in the MIMIC-III dataset \cite{johnson_mimic-iii_2016}. A large, weak entity linking dataset (of 127,150 candidate mention-UMLS pairs) was created for training. For evaluation, we annotated 1,073 mention-UMLS pairs as a gold-standard dataset. By filtering out the false positives, the proposed approach dramatically improved the precision and $F_1$ of the entity linking tool, SemEHR, with almost no loss of recall.

We further evaluated the phenotype confirmation models from discharge summaries to radiology reports in US MIMIC-III and UK NHS Tayside through either a direct transfer of the model or a weakly supervised re-training from new clinical notes. Almost perfect (100\%) recall was achieved with a dramatic absolute increase of precision by over 30\% to 50\% with re-training and parameter tuning. This demonstrates that the approach can be efficiently adapted to identify rare disease phenotypes in another type of clinical notes and from another institution. Our annotated datasets on discharge summaries and radiology reports in MIMIC-III and our implementation of the overall approach are publicly available\footnote{\url{https://github.com/acadTags/Rare-disease-identification}}.

As far as we know, this is the first study on text phenotyping of rare diseases using weak supervision, with the application on clinical notes of different types and institutions. Our findings will shed light on using weakly supervised approaches and contextual representations for text phenotyping from clinical notes. The overall approach to identifying rare disease cohorts has the potential to support epidemiology and clinical decision making for better care. 

\section*{Background and Related Work}
\label{sec:related-work}
\textbf{Text phenotyping with ontologies.} Compared to the efficient and gradually economical genotyping (i.e. sequencing genomics information), \emph{phenotyping} usually needs high-throughput computational approaches for the extraction of diseases and traits from electronic health records (EHRs) \cite{Pathak2013,Chen2013}. Clinical codes (e.g. with International Classification of Diseases, ICD) are a common source typically used regarding their ease of retrieval for phenotyping. However, ICD codes are usually less specific to define nuanced diseases or traits (e.g. rare diseases \cite{richesson2014}) and are likely to be incomplete or under-coded \cite{searle2020}, which may cause erroneous and missing cases in phenotyping. An alternative source for phenotyping is free-text clinical notes in the EHRs. It is shown in a previous systematic review of cohort identification from EHRs \cite{Ford2016} that text phenotyping (or case detection) achieves on average higher precision (or positive prediction value) and recall (or sensitivity) than code-based phenotyping, and combining both sources (texts and codes) achieved greatly improved phenotyping results. Text phenotyping also requires understanding the wider \emph{contextual features} of the matched concepts, including negation (i.e. whether negated or hypothetical), experiencer (i.e. whether experienced by the patient or someone else), and temporality (i.e. whether historical) \cite{Harkema2009context,wu2019}. These contextual features have been reasonably well detected with rule-based approaches, e.g. \cite{Harkema2009context}, and applied in Bio-YODIE and SemEHR, and more recently with neural network methods, e.g. in MedCAT \cite{Kraljevic2021}. 

Ontologies are essential for text phenotyping as they define the concepts and terms of diseases and traits. These concepts and terms are widely used to annotate clinical notes, i.e. match to text fragments or mentions \cite{kersloot_natural_2020} and to estimate rare diseases from texts \cite{kahn2017ontology}. The task to match ontology concepts (and their terms) to mentions is formally referred to as \textit{entity linking}. One main issue of entity linking is entity ambiguity, where a mention could possibly denote different concepts or terms in an ontology \cite{shen2015}. Our work aims to improve entity linking with better disambiguation using \emph{weak supervision} and \emph{contextual mention representation}.

\textbf{Weak supervision.} Weak supervision \cite{wang_clinical_2019,ratner2019} is a strategy to efficiently create a large set of noisy labelled training data in a programmatical way using various sources containing heuristics and knowledge bases. The success of applying weak supervision in clinical NLP studies depends on two aspects, \emph{data programming} and \emph{data representation}, as suggested in \cite{wang_clinical_2019}. Efficient \emph{data programming} ensures that reliable weak data can be programmatically created for supervised learning. In clinical NLP, studies use lexical or concept filtering rules to create labelled data to extract nuanced categories (e.g. suicidal ideation \cite{cusick2021} or lifestyle factors for Alzheimer's Disease \cite{shen2022classifying}) from clinical texts. We extend over this line of research by using ontologies and a medical concept labelling tool with two specific rules to create reliable weak data to extract rare diseases. The second aspect is \emph{data representation}, representing the contexts and semantics in the data into vectors in a high-dimensional space for subsequent steps in machine learning. For deep learning methods, previous studies \cite{wang_clinical_2019,shen2022classifying} proposed to use neural word embeddings and more recently using BERT \cite{devlin-etal-2019-bert} to represent the contexts of the textual data. We follow this direction to apply weak supervision with contextual representations for rare disease phenotyping.

\textbf{Contextual Representation.} The most significant, recent progress in NLP is the contextual representations pre-trained using Transformers \cite{vaswani2017attention} from a very large corpus \cite{devlin-etal-2019-bert}. The most representative contextual representation is BERT \cite{devlin-etal-2019-bert}. The pre-training task for BERT learns a masked language model with next sentence prediction, trained with a vast amount of curated texts on the Web (e.g. BookCorpus and English Wikipedia) using a 12 or 24 layered deep neural network mainly composed of multi-head self-attentions blocks. The learned parameters in the large neural network can then be applied to a wide range of downstream tasks, e.g. text classification, Named Entity Recognition, and question answering, with superior performance than the previous, task-specific models \cite{devlin-etal-2019-bert}. Contextual representations have been adapted to the clinical domain by pre-training using biomedical publications, clinical notes, and clinical ontologies. The notable models include but are not limited to BlueBERT \cite{peng2019transfer} (BERT further pre-trained with PubMed abstracts and MIMIC-III clinical notes), PubMedBERT \cite{Gu2021} (pre-trained from scratch with PubMed abstracts and full texts), SapBERT \cite{liu2021sap} (PubMedBERT further pre-trained with UMLS concepts), etc. We adapt the contextual representation methods for the mentions or text fragments to improve entity linking.

\section*{Method}
\label{sec:method}

In this section, we will describe the ontology-driven method, the weak supervision for entity linking, contextual mention representation, and model training and inferencing.

\subsection*{Entity Linking and Ontology Matching}

\textbf{Entity Linking.} Given a set of entities $E$ in an ontology and a collection of documents (e.g. clinical notes), entity linking aims to match a mention (or text fragment) $m$ to its corresponding entity $e \in E$ in the ontology \cite{shen2015}. The mention $m$ is a sequence of tokens in a document which potentially refers to one or more named entities and is usually identified in advance during the named entity recognition stage \cite{shen2015}. For Named Entity Recognition and Linking (NER+L) tools with a very large number of entities, e.g. Bio-YODIE \cite{gorrell2018}, SemEHR \cite{Wu2018semehr}, and MedCAT \cite{Kraljevic2021}, a mention $m$ is recognised at the same time when it is linked to a concept in an ontology; this is usually realised through string matching \cite{gorrell2018,Kraljevic2021}.

We applied SemEHR, a medical NER+L tool widely deployed in Trusted Research Environments (or Data Safe Havens) and servers in the UK. Previously, high recall and $F_1$ (around 90\%) were reported on sub-phenotyping with stroke from texts with SemEHR \cite{gorinski_named_2019}. The output is a set of \emph{mention-UMLS} pairs, where each mention is in a context window and with a name of the document structure (or the template section of the clinical note) if available. SemEHR adapts Bio-YODIE as its main NLP module, enhanced with a search interface and continuous learning functionalities based on users' feedback labels and rule-based and machine learning methods. Bio-YODIE can efficiently extract UMLSs from texts using a string matching based approach. When there is an ambiguous mention, time-efficient NER+L systems like Bio-YODIE mainly assume a corpus-based prior to assign the same, most frequent UMLS to the mention regardless of its context or surrounding texts \cite{gorrell2018}. This can result in many false positive phenotypes, mostly regarding the abbreviations in the clinical notes. For example in Table \ref{fp_example}, none of the identified ``HD'' mentions indicate a type of disease, according to the context. While SemEHR has a continuous learning functionality to classify and correct the errors, the approach relies on users' feedback labels and requires time from clinical experts.

\begin{table}[t]
\caption{Examples of false positives mention-UMLS pairs in entity linking identified from SemEHR and Bio-YODIE. Each mention is bolded in its context window.}
\scriptsize
\center
\label{fp_example}
\begin{tabular}{ll|p{2cm}}
\cline{1-3}
\textbf{Mention} in a context window                            & Meaning                    & \emph{False positive} UMLS \\
\cline{1-3}
 temporary \textbf{HD} line was pulled.                    & Medical device              & \multirow{4}{*}{\begin{tabular}[c]{@{}l@{}}Huntington Disease \\(C0020179) or \\Hodgkin Disease\\(C0019829)\end{tabular}} \\
... male with ESRD on \textbf{HD} ...                             & Haemodialysis    &                                                                                        \\
... Asacol \textbf{HD} 800 mg Tablet ... &  Medication &                                                                                        \\
CT scan on \textbf{HD9} showed ...                              & Hospital Day                &                                                                                        \\
\cline{1-3}
\end{tabular}
\end{table}

\textbf{Ontology Matching.} Another issue in entity linking is the variations of terms that may be missed in the process \cite{shen2015}. This can be addressed by using the rich term variations in the metathesaurus UMLS as an intermediary dictionary with \emph{ontology matching} to match concepts in UMLS to ORDO. \emph{Ontology matching} (or mapping) is the task of finding the correspondence between two ontologies \cite{noy_ontology_2009}. Each correspondence is represented as a triple $<e, e\prime, r>$, where $e$ and $e\prime$ denote an entity in the ontology $O$ and $O\prime$, respectively, and $r$ denotes a relation that holds between the two entities \cite[p.~43]{euzenat_matching_2013}. The main form of an entity in an ontology is a \textit{concept} or a class, denoted as $c \in C$ \cite[p.~34]{euzenat_matching_2013}. In ORDO, the matching of an ORDO concept to UMLS and ICD-10 concepts are available as cross references \cite{vasant2014ordo}, for example for Orphanet\_3325 (Heparin-induced thrombocytopenia), there exist correspondences $<\text{Orphanet\_3325},\text{UMLS:C0272285},\text{E}>$, where the relation E denote ``Exact matching''. We use E (Exact matching) or BTNT (ORDO's Broader Term maps to a Narrower Term) to ensure the matched term is a rare disease (and removed NTBT relations). We further added a rule (``isNotGroupOfDisorders'') to filter out the Group of Disorders, e.g. Orphanet\_181422 (Rare hyperlipidemia), which were mostly matched to a common disease in the UMLS, e.g. to C0020473 (hyperlipidemia). More details and examples of ontology matching are presented in Table S2-2 in Supplementary material 2.

\subsection*{Weak Supervision for Phenotype Confirmation Model}
To address the issue of ambiguous mentions, we propose weak supervision based on rules for labelled data creation with context mention embeddings for representation. When both data and representations are created, a classifier can be learned to decide whether a mention linked to UMLS in the context indicates a correct phenotype of the patient.

\textbf{Weakly Supervised Data Creation.} The idea in the weak data creation is to create rules that can complement the existing tool (e.g. SemEHR) to create reliable mention-UMLS pairs for training. The whole data creation process for weak supervision is described in the Algorithm \ref{WS:data_programming}. The candidate mention-UMLS pairs from an NER+L tool are denoted as a list of 5-element tuples $L$ (i.e. links), where each tuple includes a mention start position $m_{start}$, a mention end position $m_{end}$, a rare disease UMLS concept $c^{\text{rare}}_{\text{UMLS}}$, the context window of the mention $t$, and the name $s$ of the document structure where the mention is located. We propose two rules as functions on mention-UMLS pairs, \textit{mention character length} rule, $\lambda_1$, and \textit{``prevalence''} rule, $\lambda_2$, as shown in the blue blocks in Figure \ref{pipeline-weak}. Given that abbreviations (like ``HD'' in Table \ref{fp_example}) are usually ambiguous and falsely linked by the NER+L tools, the mention character length rule $\lambda_1$ satisfies when the mention has more than $l$ (default as 3) characters, i.e. $m_{end} - m_{start} > l$, otherwise as \emph{False}. Given that rare diseases usually have a very low prevalence \cite{scot_gov_2021,textoris_genetic_2014} and rare disease mentions usually have a low frequency in a consecutive sample of clinical notes, the ``prevalence'' rule $\lambda_2$ satisfies when the UMLS concept represents a very small percentage $p$ (default as 0.5\%) in the whole number of candidate links $|L|$, i.e. $\frac{\text{Freq}(c)}{|L|} < p$, otherwise as \emph{False}. This is an attempt to integrate an estimated epidemiological rule into weak supervision for text phenotyping.

\begin{algorithm}[t]
\caption{Weakly supervised data creation}

\label{WS:data_programming}
\SetAlgoLined
\KwRequire{$T$, document set; $c_{UMLS} \in O_{UMLS}$, UMLS concepts and ontology; $c_{ORDO} \in O_{ORDO}$, ORDO concepts and ontology.}
\KwEnsure{$D_{weak}$, weakly labelled data}

Initialise $D_{weak} \gets \emptyset$\;
\small{$O^{\text{rare}}_{\text{UMLS}} = \{c_{UMLS} |c_{UMLS} \in <c_{ORDO},c_{UMLS},r> \}$\;}
$L \gets \text{SemEHR}(T,O^{\text{rare}}_{\text{UMLS}})$\;
\For{each $<m_{start},m_{end},c^{\text{rare}}_{\text{UMLS}},t,s>$ in $L$}{
    $\lambda_1 = m_{end} - m_{start} > l$\;
    $\lambda_2 = \frac{\text{Freq}(c^{\text{rare}}_{\text{UMLS}})}{|L|} < p$\;
    \If {$\lambda_1 \text{ XNOR } \lambda_2$}{
        $y_{weak} \gets \lambda_1 \text{ AND } \lambda_2$\;
        $L_{weak} \gets \text{append}(L,y_{weak})$\;
        $D_{weak} \gets D_{weak} \cup L_{weak}$\;
    }
}
\end{algorithm}

The final rule-based weak labelling function $\lambda$ is defined as \emph{True} (i.e, mention-UMLS indicates a correct phenotype of the patient) when both rules $\lambda_1$ and $\lambda_2$ are satisfied, and as \emph{False} when both rules are not satisfied. The data selection is equivalent to an XNOR logic operator (selected if and only if both rules are True or both are False) and the data labelling is equivalent to an AND operator of the rules. This ensures that only data that are consistently checked by both rules are weakly labelled. The binary weak label, $y_{weak} \in \{0,1\}$, is then appended to each mention-UMLS pair to create the weakly labelled data $D_{weak}$.

The mention length threshold $l$ and the ``prevalence'' threshold $p$ are selected to ensure a sufficient amount of reliable, weak data generated. We empirically determine the best values of $l$ (as 3 or 4) and $p$ (as 0.005 or 0.01) based on the validation set or a small number of annotated data solely for evaluation (results on MIMIC-III discharge summaries in Table S1-1 in the Supplementary material 1).

\textbf{Contextual Mention Representation.} We use a clinically pre-trained BERT model (e.g. BlueBERT, as described in the related work) to represent the mention in its context window $t$ in the weakly labelled data $D_{weak}$. A BERT model can be succinctly described as the Equations \ref{BERT}. We excluded layer normalisation, dropout, and other functions and parameters in the equations for simplicity. The output $H^n \in R^{|\text{tokens}|,d}$ is a matrix that can be used as the layer for the subsequent task, where $|\text{tokens}|$ is the length of sequence after tokenisation and $d$ denotes the dimensionality (usually 768 for BERTnorm and 1024 for BERTlarge). \text{FFNN}() is a feed-forward neural network of two linear transformations with a ReLU activation function in between, and \text{MultiHead}() is a multi-head self-attention layer that models multiple forms of alignment from the tokens to themselves; and the three inputs represent matrices of queries ($Q$), keys ($K$), and values ($V$), respectively, linearly transformed from $H^i$. We refer readers for the details of the Transformers and BERT architectures to \cite{vaswani2017attention,devlin-etal-2019-bert}.
\begin{equation}\label{BERT}
\begin{split}
  H^{i+1} & = \text{FFNN}(\text{MultiHead}(W_QH^i,W_KH^i,W_VH^i)) \\
  H^0 & = \text{Embedding}(\text{Tokenize}(t))
\end{split}
\end{equation}

The contextual understanding mainly comes from self-attention (as $\text{softmax}(\frac{QK^{T}}{\sqrt{d_k}})V$, where $d_k$ is a scaling factor) that captures the importance of every other token to each token. These parameters have been pre-trained based on massive corpora from general and medical domains. The hidden layers in BERT, $H$ can be used as \textit{static} embeddings to represent a sequence. We extract the second-last layer $H^{n-1}$ in BERT as static embedding (or features) for the subsequent task, according to the results that $H^{n-1}$ has the best feature-based results among any single layers in $H$ for an NER task \cite{devlin-etal-2019-bert}. A plausible explanation for this is that the last layer is more biased towards the training loss (e.g. masked language model and next sentence prediction), while the second-to-last layer better represents the contextual information of the sentence.

The selection of the specific BERT model generally favours models pre-trained with in-domain (i.e. clinical) corpora \cite{gururangan2020} and is empirically based on results (e.g. $F_1$ scores) on the validation set. We will compare and analyse different BERT models in the experiments (see Table \ref{umls_linking_embedding_results}).

The overall weak supervision data representation and model training process is described in Algorithm \ref{WS:data_rep_train}. We use $H^{n-1} \gets \text{BERT}(t)$ to denote the whole process above. Mean pooling, as empirically suggested in \cite{ma2019universal}, is applied to create a final vector $v$. We define a contextual mention representation where only the tokens within the mention are included, i.e. $v \gets \text{mean}(H^{n-1}[m^{\text{token}}_{start},m^{\text{token}}_{end}])$. The start and end tokens' position of the mention $m^{\text{token}}_{start}$ and $m^{\text{token}}_{end}$ are derived based on the WordPiece tokenizer of the BERT model and the original position of the mention.

We also experimented with two encoding strategies, mention masking and using document structure name $s$ (see line 3 in Algorithm \ref{WS:data_rep_train}), that allow a more flexible representation of the contexts. Non-masked encoding with document structures provided better results on the validation set (see Table S1-2 in Supplementary material 1).

\textbf{Model Training and Inference.} Finally, a \emph{phenotype confirmation model} can be trained from the weakly labelled data. The contextual mention representation $v$, as static embedding, is fed into a binary classification model. We use logistic regression as the training model (in Train\_and\_validate() in Algorithm \ref{WS:data_rep_train}), which is similar to adding a feed-forward layer on top of the static pre-trained layer in BERT with sigmoid activation. We also compared this static embedding approach to fine-tuning the whole BERT model in the experiments.

\begin{algorithm}[t]
\caption{Weakly supervised data representation and model training}

\label{WS:data_rep_train}
\SetAlgoLined
\KwRequire{$<m_{start},m_{end},c^{\text{rare}}_{\text{UMLS}},t,s,y_{weak}> \in D_{weak}$}
\KwEnsure{$M_{weak}$, the phenotype confirmation model}

Initialise $X_{weak} \gets \emptyset$, $Y_{weak} \gets \emptyset$\;
\For{each $<m_{start},m_{end},c^{\text{rare}}_{\text{UMLS}},t,s,y_{weak}>$ in $D_{weak}$}{
    $t \gets \text{concatenate}(t,\text{[SEP]},s)$, if $s$ is not \textit{null}\;
    $H^{n-1} \gets \text{BERT}(t)$\;
    $m^{\text{token}}_{start}, m^{\text{token}}_{end} \gets \text{tokenize}(t,m_{start},m_{end})$\;
    $v \gets \text{mean}(H^{n-1}[m^{\text{token}}_{start},m^{\text{token}}_{end}])$\;
    $X_{weak} \gets X_{weak} \cup v$\;
    $Y_{weak} \gets Y_{weak} \cup y_{weak}$\;
}

$M_{weak} \gets \text{Train\_and\_Validate}(X_{weak},Y_{weak})$\;
\end{algorithm}

The inference stage is succinctly defined in Equation \ref{inference}. We use SemEHR to extract candidate mention-UMLS pairs from a clinical note $d$. We then transform each instance into a contextual mention representation (see line 3-6 in Algorithm \ref{WS:data_rep_train}), denoted as the function $V_\text{BERT}()$. After selecting the patients' phenotype in $O^{\text{rare}}_{\text{UMLS}}$ with $M_{weak}$, we can then use the correspondence between UMLS and ORDO, denoted as $\text{OM}_{U\rightarrow O}$, to obtain the final set of rare disease phenotypes $C^{d}_{ORDO}$ as concepts in ORDO.

\begin{equation}\label{inference}
\scriptsize
  C^{d}_{ORDO} = \text{OM}_{U\rightarrow O}(M_{weak}(V_\text{BERT}(\text{SemEHR}(d,O^{rare}_{UMLS})))))
\end{equation}

\section*{Experiments}
\label{sec:experiments}

We evaluated the above ontology-driven and weakly supervised algorithms on MIMIC-III discharge summaries and further validated the approach with MIMIC-III radiology reports and NHS Tayside brain imaging reports. For validation and testing, we manually annotated a small number of mention-to-UMLS pairs from each of the datasets. We present results on each part of the system, Text-to-UMLS and UMLS-to-ORDO. For Text-to-UMLS, we carried out extensive experiments to study the best combination of parameters in weak labelling rules, the encoding strategies, with a comparison between weak and strong supervision. We then show the whole pipeline can support rare disease phenotyping by enriching the traditional method using ICD codes. Finally, we show that the proposed approach can easily generalise or be adapted to a new type of clinical note, radiology reports, in the same or another institution.

\subsection*{Data Processing and Annotation}
We evaluated the proposed NLP pipeline with three datasets in two healthcare institutions in the US and the UK. The main dataset we used was the discharge summaries (n=59,652) in MIMIC-III (``Medical Information Mart for Intensive Care'') dataset \cite{johnson_mimic-iii_2016}, which contains clinical data from adult patients admitted to the ICU in the Beth Israel Deaconess Medical Center in Boston, Massachusetts between 2001 and 2012. We were granted access to MIMIC-III through PhysioNet after completing the ethical training by the Collaborative Institutional Training Initiative program. MIMIC-III data are supposed to contain rich rare disease mentions, as a large number of rare diseases (especially genetic disorders) can lead to an ICU (intensive care unit) admission \cite{textoris_genetic_2014}.

The manual ICD-9 codes (i.e. ICD-9-CM) of the MIMIC-III admissions allow us to compare code-based phenotyping with text phenotyping for rare diseases. We linked ICD-9 codes to ICD-10 codes using the matching from the Ministry of Health, New Zealand \cite{icd9to10newzeland} and linked ICD-9 to UMLS codes based on the ICD-9 ontology in BioPortal \cite{icd9ontology}, as shown in Figure \ref{pipeline-main}. We used ORDO version 3.0 (released 07/03/2020), which contained 14,501 concepts or classes related to rare diseases. We selected the ORDO concepts which have linkage to UMLS and ICD-10 in this study as this supports the interoperability (e.g. linking and traversing) among the clinical terminologies; this resulted in a set of 4,064 rare disease concepts\footnote{The most 5 frequent UMLS (version 2020AB) semantic types of the 4,064 linked ORDO concepts: T047 (Disease or Syndrome, 3,245 concepts,  79.8\%), T019 (Congenital Abnormality, 465 concepts, 11.4\%), T191 (Neoplastic Process, 374 concepts, 9.2\%), T049 (Cell or Molecular Dysfunction, 35 concepts, 0.9\%), and T046 (Pathologic Function, 19 concepts, 0.5\%); note that 160 concepts (3.9\%) are associated with two semantic types.}. We focus on this essential set of overlapped rare diseases and the coverage is improving as the mappings are being updated; we leave the ORDO concepts without both ICD-10 and UMLS linkage for future research.

After processing the discharge summaries with a SemEHR database instance\footnote{\url{https://github.com/CogStack/CogStack-SemEHR}} \cite{Wu2018semehr} with rule-based contextual filtering on negation and experiencer based on \cite{Harkema2009context}, we obtained 127,150 candidate mention-UMLS pairs for the UMLS concepts linked to ORDO. After applying the weak labelling function with the two rules, we finally obtained 15,598 positive and 74,217 negative data, and 37,335 non-labelled data or mention-UMLS pairs.

We further applied the same preprocessing steps with the MIMIC-III radiology reports (n=522,279) and NHS Tayside brain imaging reports (n=156,618). MIMIC-III radiology reports are from the same institution and within the same time span as in MIMIC-III discharge summaries \cite{johnson_mimic-iii_2016}. The Tayside data contain the routine brain MRI and CT scans from the National Health Service (NHS) Tayside Health Board, which have been applied in previous NLP research \cite{gorinski_named_2019,sykes2021}. We have received NHS Tayside Caldicott Guardian approval to use the anonymised brain imaging reports for this work. 

The statistics of the three datasets, MIMIC-III discharge summaries (``Disch''), MIMIC-III radiology reports (``Rad''), and NHS Tayside brain imaging reports (``Tayside Brain Img''), with the Natural Language Processing pipeline and manual annotations, are presented in Table \ref{data-statistics}. MIMIC-III discharge summaries have \emph{proportionally} more documents associated with at least one candidate rare diseases (identified by SemEHR), quantified by $\frac{|T_{RD}|}{|D|}$: 3.4 times more than MIMIC-III radiology reports and 13.3 times more than brain imaging reports in Tayside.

\begin{table}[ht]
	\caption{Statistics of Clinical Note Datasets with the Natural Language Processing Pipeline and Manual Annotations}
	\center
	\scriptsize
	\label{data-statistics}
	\begin{threeparttable}
		\begin{tabular}{llll}
			\cline{1-4}
			& MIMIC-III Disch       & MIMIC-III Rad      & Tayside Brain Img          \\
			\cline{1-4}
			$|T|$                   & 59,652              & 522,279           & 156,618           \\
			$|D|$                   & 127,150             & 109,096           & 7,761               \\
			$|D_{weak^+}|$           & 15,598              & 13,907            & 1,137             \\
			$|D_{weak^-}|$           & 74,217              & 65,171            & 2,898             \\
			$|T_{RD}|$     & 37,110              & 73,589            & 7,321                      \\
			$|T^{weak}_{RD}|$       & 10,568              & 21,102            & 2,855             \\
			\hline\hline
			$|T^{ann}|$             & 500                 & 1,000             & 5,000             \\
			$|D^{ann}|$             & 1,073               & 198               & 279+4             \\
			$|T^{ann}_{RD}|$        & 312                 & 145               & 273               \\
			\cline{1-4}
		\end{tabular}
		\begin{tablenotes}
			\item $|T|$, number of documents; $|D|$, number of mention-UMLS pairs; $|D_{weak^+}|$, $|D_{weak^-}|$, number of weakly labelled positive and negative mention-UMLS pairs, respectively; $|T_{RD}|$, $|T^{weak}_{RD}|$, number of documents associated with one or more rare diseases detected by SemEHR and SemEHR+WS (i.e. further with weak supervision), respectively; $|T^{ann}|$, $|D^{ann}|$, $|T^{ann}_{RD}|$, number of documents sampled, number of mention-UMLS pairs sampled, and number of the sampled documents with one or more rare diseases identified by SemEHR, respectively. For Tayside data, 4 new positive mention-UMLS pairs in $|D_{ann}|$ were identified from the reports during the manual annotation.
		\end{tablenotes}
	\end{threeparttable}
\end{table}

\textbf{Data Annotation.} For evaluation, we created a gold standard dataset of 1,073 candidate mention-UMLS-ORDO triplets (with each mention in a context window) generated by SemEHR and ontology matching in ORDO, from a set of 500 randomly sampled discharge summaries from MIMIC-III, of which 312 (or 62.5\%) discharge summaries have at least one \textit{candidate} or potential ``rare disease'' mention. There were in total 95 types of rare disease associated with the mentions. Annotators were asked to label whether a mention-UMLS pair truly indicates a phenotype of the patient with an annotation guideline of detailed examples on hypothetical mentions. The mention-UMLS pairs were annotated by 3 domain experts, including two research fellows and one PhD student in Medical Informatics (MI). Based on the random 200 mention-UMLS pairs annotated by all 3 domain experts, the multi-rater Kappa value was 0.76. ORDO-to-UMLS concept matching was annotated by 2 domain experts (a research fellow and a PhD student in MI) and obtained a Kappa of 0.72. All contradictory and unsure annotations were resolved by a research fellow in biomedical science and MI. We used the first 400 data instances for model validation and the rest 673 for final testing.

To study how the model performs when it is directly transferred to or re-trained on other clinical notes, we further annotated 198 candidate mention-UMLS pairs in a sample of 1,000 radiology reports in MIMIC-III \cite{johnson_mimic-iii_2016} and 279 candidate mention-UMLS pairs (with 4 new manually identified mentions) in a sample of 5,000 brain imaging reports in NHS Tayside \cite{gorinski_named_2019}. Each dataset was annotated by two researchers in clinical science or MI with contradictions addressed by another researcher. The Kappa for MIMIC-III radiology reports and NHS Tayside reports were 0.88 and 0.86, respectively. 

To note that the evaluation set is independent of the rules used for weak supervision, thus abbreviations and ``popular'' disease mentions were in the validation and testing data. This helps to test whether the phenotype confirmation model trained on the rule-based weakly labelled data can generalise to the full scenario that also contains the \textit{unseen} mentions, which were filtered out during weak supervision.

\subsection*{Implementation Details}
We used the open-source tool, bert-as-service\footnote{\url{https://bert-as-service.readthedocs.io/en/latest/}} \cite{xiao2019bertservice}, built on Google AI's BERT implementation with Python Tensorflow\footnote{\url{https://github.com/google-research/bert}} \cite{devlin-etal-2019-bert} for contextual mention representation. We tested a range of pre-trained BERT models (BERT, BlueBERT, PubMedBERT, and SapBERT) and selected BlueBERT-base \cite{peng2019transfer} based on results on the validation set (see Table \ref{umls_linking_embedding_results}). We then trained a logistic regression model with the representations, with default configuration using scikit-learn \cite{scikit-learn} on the weakly labelled mention-UMLS pairs. We also implemented a word2vec embedding baseline with Gensim\footnote{\url{https://radimrehurek.com/gensim/models/word2vec.html}} and a BERT fine-tuning baseline with Huggingface Transformers\footnote{\url{https://github.com/huggingface/transformers}}, with detailed parameters in \textit{Embedding and Fine-tuning Settings} in Supplementary material 1. Our implementation of the experiments is available at \url{https://github.com/acadTags/Rare-disease-identification}.

As baselines, we compared the proposed approach (``SemEHR+WS'') with SemEHR with the two rules only using an OR operation for the interest of higher recall (``SemEHR+rules''). We evaluated the baselines using precision, recall, and $F_1$ scores. Note that SemEHR had a reference recall of 100\% as all candidate ``rare disease'' mentions were identified by SemEHR, which was the starting source for the annotations.\footnote{We also benchmarked the performance on MIMIC-III discharge summaries with recent NER+L tools, MedCAT \cite{Kraljevic2021} and Google Healthcare Natural Language API \cite{Bodnari2020}. However, given that the results (especially recall) may favour SemEHR-based methods, we do not formally report the results of the two NER+L tools but make them available at \url{https://github.com/acadTags/Rare-disease-identification/blob/main/supp-results}.}

We tuned the two parameters $l$ and $p$ (to 3 and 0.5\%, respectively, if not specified) in the weak labelling rules (in Algorithm \ref{WS:data_programming}) by grid search based on the performance of validation data in MIMIC-III discharge summaries. The detailed parameter tuning results of $l$ and $p$ are in Table S1-1 in Supplementary material 1, \textit{Weak Rule Parameter Tuning}. We also tuned the size of context windows (default as 5), which however, did not affect the performance, probably because our final representation was based on the position of the mention in the BERT layer (see line 6 in Algorithm \ref{WS:data_rep_train}). Also, we tuned the optimal number of random training mention-UMLS pairs needed (n=9k) based on the validation set, which had little impact on the results ($<$1\% $F_1$ score).

In contrast to weak supervision (WS), we also provide results on strong supervision (SS), the traditional approach that trains a model from full manually labelled data. For MIMIC-III discharge summaries, we used the first 400 validation set in the full 1,073 mentions to train a model, $M_{strong}$, and test on the rest 673 mentions with the same inferencing step in Equation \ref{inference} but using $M_{strong}$ instead of $M_{weak}$. As manually labelled data are usually more reliable than weakly labelled data, the performance of strong supervision is considered as an upper bound in studies in weak supervision \cite{fries2021ontology,ratner2020snorkel}.

We provide the results regarding each step in the pipeline (in Figure \ref{pipeline-main}), Text-to-UMLS linking and UMLS-to-ORDO matching, followed by the overall results on rare disease identification, Text-to-ORDO linking and admission-level ORDO concept prediction.

\subsection*{Main Results: Text-to-UMLS linking}
\begin{table*}[th]
\caption{Evaluation results of Text-to-UMLS linking on validation and testing data from MIMIC-III discharge summaries}
\scriptsize
\center
\label{umls_linking_results}
\begin{threeparttable}
\begin{tabular}{llll|lll||lll|lll}
\cline{1-13}
                            & \multicolumn{3}{l}{validation (n=142+/400)}         & \multicolumn{3}{l}{test (n=187+/673)}               & \multicolumn{3}{l}{\begin{tabular}[c]{@{}l@{}}test, \emph{seen} in WS (n=80+/499)\\ i.e. both rules [not] satisfied\end{tabular}} & \multicolumn{3}{l}{\begin{tabular}[c]{@{}l@{}}test, \emph{unseen} in WS (n=107+/174)\\i.e. only one rule satisfied \end{tabular}} \\
Text to UMLS                & P             & R              & $F_1$         & P             & R              & $F_1$         & P                  & R                   & $F_1$             & P                  & R                   & $F_1$             \\
\cline{1-13}
SemEHR \cite{Wu2018semehr} & 35.5          & \textbf{100.0} & 52.4          & 27.8          & \textbf{100.0} & 43.5          & 16.0               & \textbf{100.0}      & 27.6 & 61.5               & \textbf{100.0}      & 76.2                            \\
+ rules                     & 80.9          & 89.4           & 84.9          & 68.6          & 94.7           & 79.6          & \textbf{83.3}      & 87.5                & \textbf{85.4} & 61.5               & \textbf{100.0}      & 76.2                   \\
+ WS (rules+BERT)              & \textbf{92.0} & 89.4           & \textbf{90.7} & \textbf{81.4} & 91.4           & \textbf{86.1}  & \textbf{83.3}      & 87.5                & \textbf{85.4} & 80.2               & 94.4       & 86.7                   \\
\hline\hline
+ SS (anns+BERT) & - & -           & - & \textbf{88.4} & 93.6           & \textbf{90.9}      & \textbf{87.7} & 88.8 & \textbf{88.2} & 88.9 & 97.2 & \textbf{92.9} \\
\cline{1-13}
\end{tabular}
\begin{tablenotes}
\item The column statistics (n=$N_+$+/$N$) show the number of positive data $N_+$ and all samples $N$ in the dataset. SemEHR has a perfect reference recall, because all candidate mention-UMLS pairs were created using the tool. WS, weak supervision; SS, strong supervision. BlueBERT-base (PubMed+MIMIC-III) was used as the BERT model. The best scores, either or not considering strong supervision (SS), are bolded.
\end{tablenotes}
\end{threeparttable}
\end{table*}

Table \ref{umls_linking_results} shows the validation and testing results of Text-to-UMLS linking. With weak supervision (WS), the precision and $F_1$ of SemEHR has been greatly improved by around 55\% and 40\% absolute value, respectively, for both validation and testing data. Adding the two customised rules already improved the testing performance greatly by over 30\% $F_1$ to SemEHR (as shown in SemEHR+rules), which validates the efficiency of the two proposed rules with the NER+L tool to create reliable weak annotations. Adding WS further outperformed the SemEHR+rules setting absolutely by around 10\% precision (and 5\% $F_1$), showing the usefulness of the contextual mention representation on filtering out false positives. The recall dropped slightly after introducing the two rules. This indicates the bias or noise in the rules with the current threshold ($p$ as 0.5\% and $l$ as 3). Results with weak supervision are within a small gap of 5\% $F_1$ of strong supervision with hand-labelled data. This, overall, demonstrates the potential of WS to improve text phenotype entity linking.

As a solid evaluation needs to assess the system with different biased test sets, we further split the testing data into those weakly labelled or unlabelled during the weak supervision. This helps analyse the impact of the rule-based weak supervision on the testing performance. ``Seen'' data mean that the mention-UMLS pairs were weakly labelled with $\lambda$, i.e. with both rules satisfied or both not satisfied (see line 7-11 in Algorithm \ref{WS:data_programming}); ``unseen'' data mean that only one of the rules was satisfied so that the data were not labelled in the process. WS improved the performance of SemEHR in both settings: while the weakly ``seen'' data were dramatically boosted by rules (by nearly 50\% $F_1$), the ``unseen'' data were greatly improved (by 10\% $F_1$) through the model generalised with contextual representations. 

The ``unseen'' data can be further split into the case that only the mention character length rule ($\lambda_1$) or the prevalence rule ($\lambda_2$) is satisfied. The former, ``unseen-$\lambda_1$'' testing set (n=127, where 96 are positive mentions) has more mentions than the latter, ``unseen-$\lambda_2$'' (n=47, where 11 are positive mentions). SemEHR+WS obtained substantially better P/R/$F_1$ performance on ``unseen-$\lambda_1$'' (84.1/99.0/90.9) than ``unseen-$\lambda_2$'' (46.2/54.5/50.0). This shows that mentions that are infrequent abbreviations (i.e., ``unseen-$\lambda_2$'') tend to be more challenging than frequent non-abbreviations (i.e., ``unseen-$\lambda_1$''). In both scenarios, SemEHR+WS performed the best $F_1$ among the baselines except for strong supervision (SemEHR+SS). However, given that the number of testing samples is small, e.g. only 11 positive mentions for ``unseen-$\lambda_2$'', we do not formally report the breakdown of results to draw solid conclusions.

\textbf{Embedding and Encoding Strategies.} We compared the different \emph{embedding} methods, including word embeddings and several BERT models pre-trained from different sources. Table \ref{umls_linking_embedding_results} shows that contextual mention embeddings (e.g. with BERT, described in lines 4-6 in Algorithm \ref{WS:data_rep_train}) based methods greatly outperformed word embeddings, although increasing the dimensionality of word2vec embeddings improved their recall and $F_1$. For the contextual mention embeddings, we compared the vanilla BERT and representative pre-trained BERT models in the biomedical domain. We observed that BlueBERT, pre-trained using the in-domain (or same-data), MIMIC-III clinical notes, outperformed the various BERT models only from general domains (e.g. BERT), biomedical publications (e.g. PubMedBERT), or clinical ontologies (e.g. SapBERT). This supports the use of in-domain pre-trained models, e.g. BlueBERT for the task, corroborating the conclusion from \cite{gururangan2020}. We also see that \textit{neither} using fine-tuning (cf. feature-based) nor the large version of BlueBERT could improve the performance, which is probably because they introduce more learnable parameters (and a larger model size for BlueBERT-large), thus likely overfitting the weakly labelled data and underperforming on the real, testing data. We further compare the \emph{encoding} strategies and found that non-masked encoding (with document structures) achieved the best $F_1$ scores on the validation data (see Table S1-2 in Supplementary material 1).

\begin{table}[t]
\caption{Comparison among embeddings for weakly supervised Text-to-UMLS linking from MIMIC-III discharge summaries}
\scriptsize
\center
\label{umls_linking_embedding_results}
\begin{threeparttable}
\begin{tabular}{llll|lll}
\cline{1-7}
                            & \multicolumn{3}{l}{validation (n=142+/400)}         & \multicolumn{3}{l}{test (n=187+/673)}               \\
Text to UMLS      & P             & R              & $F_1$         & P             & R              & $F_1$       \\
\cline{1-7}
Word2Vec-100                     & 86.6          & 50.0           & 63.4          & 85.1          & 61.0           & 71.0\\
Word2Vec-300                     & 85.7          & 59.2           & 70.0          & 80.7          & 69.5           & 74.7\\
Word2Vec-768                     & 85.1          & 68.3           & 75.8          & 78.9          & 78.1           & 78.5\\
\cline{1-7}
BERT    & 88.1          & 83.8           & 85.9          & 79.5         & 91.4           & 85.1            \\
PubMedBERT & 88.7          & 77.5           & 82.7          & 79.6          & 87.7           & 83.5            \\
SapBERT & 88.3          & 79.6 & 83.7          & \textbf{80.8}          & 89.8 & 85.1         \\
BlueBERT-base & \textbf{90.1}          & \textbf{89.4}           & \textbf{89.8}          & 80.4          & \textbf{92.0}           & \textbf{85.8}                   \\
+ fine-tuning & 84.6          & 88.7           & 86.6          & 73.5          & \textbf{92.0}           & 81.7                   \\
BlueBERT-large & 89.1          & 80.3           & 84.4          & 79.0          & 88.8           & 83.6                   \\
\cline{1-7}
\end{tabular}
\begin{tablenotes}
\item The column statistics (n=$N_+$+/$N$) show number of positive data $N_+$ and all samples $N$ in the dataset. All word2vec-$k$ embeddings were pre-trained from MIMIC-III discharge summaries, representing the mention as the averaged $k$-dimensional embedding of tokens in the context window. BERT models were used as static features (in the second-last layer) if not specified with ``fine-tuning''. The best scores, either or not considering strong supervision (SS), are bolded. We did not tune the optimal number of random weakly supervised training data for BlueBERT-base model (and all other models), thus its results were slightly below those in Table \ref{umls_linking_results}.
\end{tablenotes}
\end{threeparttable}
\end{table}

\subsection*{UMLS-to-ORDO Matching Results}

For UMLS-to-ORDO ontology matching, the original accuracy by the ORDO ontology was 87.4\% (=83/95), if considering the repeated mentions in the whole 1073 evaluation data, the linking accuracy was 81.6\% (=876/1073). The most frequent three false UMLS-to-ORDO mappings in ORDO were Hyperlipidemia (C0020473) to Rare hyperlipidemia (Orphanet\_181422), Epilepsy (C0014544) to Rare epilepsy (Orphanet\_101998), and Dyslipidemias (C0242339) to Rare dyslipidemia (Orphanet\_101953), all linking a broader, common disease concept to its specific types in rare diseases under the phenome type or the upper class \cite{vasant2014ordo} of \textit{group of disorders} (Orphanet\_557492). By filtering with ORDO's phenome type using ``isNotGroupOfDisorders'' (i.e. not under \textit{group of disorders}), the UMLS-to-ORDO concept linking accuracy of the unique and repeated mentions was improved to 88.4\% (from 87.4\%) and 94.4\% (from 81.6\%), respectively, from the whole validation and testing data in the MIMIC-III discharge summaries.

\subsection*{Overall Mention-level and Admission-level Results}
\label{subsec:overall_men_adm_results}
We finally obtained the mention-level results (Text-to-ORDO) based on the two parts of the system. The results, shown in Table \ref{mention_rd_id_results}, are consistent with Text-to-UMLS results. The overall metrics are lower than Text-to-UMLS results (71.7\% vs 86.1\% for testing $F_1$ score for WS) due to the imperfect matching between UMLS and ORDO. For a perfect UMLS-to-ORDO matching, the results of the Text-to-UMLS and Text-to-ORDO should be the same.

\begin{table}[t]
\caption{Results on rare disease identification (Text-to-ORDO) from MIMIC-III discharge summaries}
\scriptsize
\center
\label{mention_rd_id_results}
\begin{threeparttable}
\begin{tabular}{llll|lll}
\cline{1-7}
                            & \multicolumn{3}{l}{validation (n=64+/400)}         & \multicolumn{3}{l}{test (n=82+/673)}               \\
Text to ORDO      & P             & R              & $F_1$         & P             & R              & $F_1$       \\
\cline{1-7}
SemEHR \cite{Wu2018semehr} & 18.7          & 95.3           & 31.3          & 13.9          & \textbf{92.7} & 24.1         \\
+ rules                     & 53.9          & 75.0           & 62.7          & 49.0          & 86.6           & 62.6               \\
+ WS (rules+BERT)           & \textbf{67.6} & 75.0           & \textbf{71.1} & \textbf{64.7} & 80.5           & \textbf{71.7}     \\
\hline\hline
+ SS (anns+BERT)            & -             & -              & -             & \textbf{73.3} & 80.5           & \textbf{76.7}      \\
\cline{1-7}
\end{tabular}
\begin{tablenotes}
\item The column statistics (n=$N_+$+/$N$) shows number of positive data $N_+$ and all samples $N$ in the dataset. WS, weak supervision; SS, strong supervision; anns, annotations. BlueBERT-base (PubMed+MIMIC-III) was used as the BERT model. The best scores, either or not considering strong supervision (SS), are bolded.
\end{tablenotes}
\end{threeparttable}
\end{table}

In the interest of detection of rare disease cases in admissions, we aggregated the mention-level results to admission-level results, where one admission may be associated with several unique rare diseases (each as a concept in ORDO). Thus, we report the standard micro-level label-based metrics for multi-label classification \cite{Gibaja2015}. Micro-level metrics count each admission to a single ORDO concept as an instance and create a confusion matrix to calculate the precision, recall, and $F_1$ scores. We were also able to obtain ICD-based results purely based on ontology matching (from ICD-9 codes to ICD-10 or UMLS concepts then finally to ORDO concepts, as shown in Figure \ref{pipeline-main}). Admission-level results were generally consistent with mention-level (Text-to-UMLS and Text-to-ORDO) results. In terms of precision and $F_1$ score, weak supervision greatly improved the performance of SemEHR and outperformed other third-party tools, slightly below strong supervision, while the recall was the same for both WS and SS. We also obtained the admission-level results of ICD codes.

Admission-level results are presented in Table S1-3 in Supplementary Material 1. It is discovered that our NLP-based approach (SemEHR+WS) achieved better precision and $F_1$ scores than the code-based approach (ICD). In terms of recall, ICD codes could only identify a few more rare diseases cases than SemEHR with weak supervision (e.g. 21 vs 20 out of 30 in the validation set and 36 vs 33 out of 42 in the test set, between ICD $\cup$ SemEHR+WS and SemEHR+WS). Note that this result may not be accurate as our annotation is based on the string matching based NER+L results from SemEHR, so the false positives from ICD-based cohorts may actually be true cases. Also, the number of positive data is much lower in admission-level results than in the mention-level (e.g. for testing data, 42 admissions vs. 187 mention-UMLS pairs). But nevertheless, our results show the essential role of free-texts and NLP methods for rare disease phenotyping; the results are consistent with the conclusion in \cite{Ford2016} regarding general diseases.

\subsection*{Error Analysis}
\label{subsec:error_analysis}
\begin{figure}[t]
  \centering
  \includegraphics[width=0.45\textwidth]{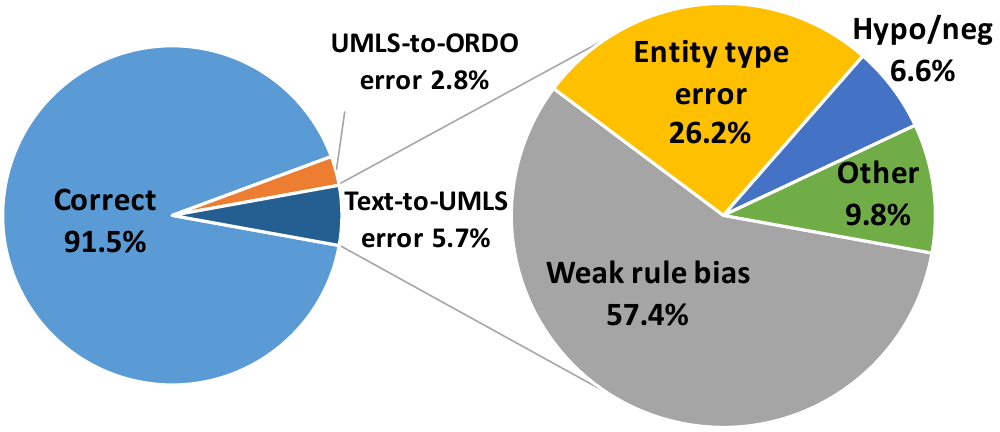}
  \caption{Error breakdown of Text-to-ORDO identification of 1,073 candidate mentions in MIMIC-III discharge summaries (Hypo/neg: Hypothetical or negation)}\label{error_pie_chart}
\end{figure}

We breakdown the errors of the proposed approach (``SemEHR+WS'') regarding Text-to-ORDO in MIMIC-III discharge summaries (see results in Table \ref{mention_rd_id_results}) in Figure \ref{error_pie_chart}. There were altogether 91 errors (including 59 false positives and 32 false negatives), representing 8.5\% from the 1,073 candidate mentions-UMLS-ORDO triplets, where 61 (or 5.7\%) were from Text-to-UMLS stage and 30 (or 2.8\%) \emph{only} from the UMLS-to-ORDO stage (and 4 in both stages).

While rules are effective for WS, they may also introduce some bias. Over half 57.4\% (or 35 of 61 errors) from the Text-to-UMLS side were likely due to the bias introduced from the weak rules, where the prediction was wrong when using the weak rules only. The other two main errors were either (i) semantic type errors (representing 26.2\% or 16 out of 61), where the mention was a (negative) laboratory test (e.g. ``legionella'') or other unrelated types (e.g. ``ENDO'' as department name) instead of a disease, or (ii) diseases of hypothetical or negative contexts (represented 6.6\% or 4 out of 61), which were not filtered out by the NER+L tool, SemEHR, and were also challenging for the annotators. The other errors (9.8\%, 6 out of 61) were due to not enough information for human to decide or no exact reason found for the error. The issues above may be addressed by combining WS with human-in-the-loop machine learning \cite{monarch2021} with adaptive rules to improve the performance. The wrong UMLS-to-ORDO ontology mappings were due to the simple heuristic (``isNotGroupOfDisorders'') which also filtered out correct mappings - this may be addressed when the official ontology matching is updated or by using a machine learning based system to correct the matching.

\subsection*{NLP vs. ICD for Rare Disease Phenotyping} 
\label{subsec:nlp_vs_icd}
We applied the trained model and the whole pipeline to process all MIMIC-III discharge summaries (n=59,652) and compared the rare disease admissions identified from NLP and ICD. The NLP approach is the proposed ontology-driven and weakly supervised pipeline. For the ICD-based results, we combined the ICD-9 codes matched to either the UMLS or ICD-10 codes linked to ORDO (see Figure \ref{pipeline-main}).

\begin{figure*}[t]
  \centering
  \includegraphics[width=0.8\textwidth]{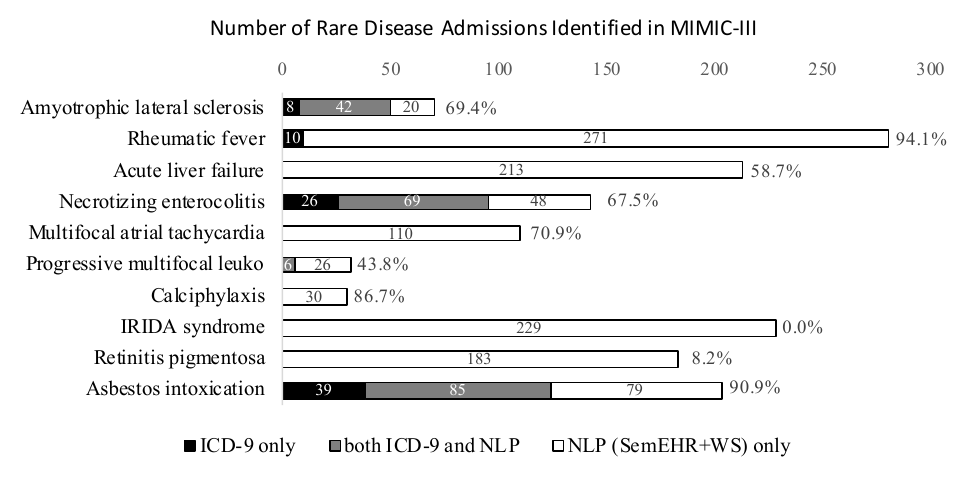}
  \caption{Number of rare disease patient stays from MIMIC-III (n=59,652): ICD (code-based) vs. NLP (text-based, with \textit{weak} supervision), for 10 selected diseases. Admissions are split into those \textit{only} identified through links from ICD-9 codes (in black), those \textit{only} identified from free texts with weak supervision (NLP, in white), and the intersection of cases from both ICD-9 and NLP (in grey). The percentage after each horizontal bar shows the accuracy of NLP based on the manual assessment of the identified cases.}\label{patient_icd_vs_nlp}
\end{figure*}

Using our NLP-based pipeline, it is possible to greatly enrich the rare disease cases identified solely from ICD codes. For most (97.2\%=453/466) types of the rare diseases, our approach mining free texts could enrich at least one (and usually many) potential rare disease case compared to the ICD-based approach. The results can be useful to identify potential cases for an alerting system for clinical care or a base for further refinement. Figure \ref{patient_icd_vs_nlp} shows the selected 10 rare diseases which were best predicted in the annotated 312 discharge summaries, however, since the support value was few (between 1 to 5) for each of the diseases in the admission-level evaluation, the results did not represent the predictions of the full 59k admission cases in MIMIC-III.

We thus further performed an extra manual evaluation to verify whether the rare disease cases identified by NLP were true phenotypes (or represented a current or past rare disease of the patient), as there was no gold reference standard. Five researchers (one in clinical science, one in biomedical science, and the remaining three in MI) screened the 1,428 cases or patient stays identified by NLP (WS or SS) regarding the 10 selected diseases, according to the definitions of the rare diseases in ORDO. The accuracy scores (the fraction of correct rare disease cases in all identified cases) of the weakly-supervised NLP-identified rare diseases are displayed after each horizontal bar in Figure \ref{patient_icd_vs_nlp}. We can see that NLP identified most rare diseases (6/10) with an accuracy score from around 70\% to over 90\%. For rheumatic fever, over 90\% of the cases were true positives, except for a few hypothetical mentions or the subject being the patient's relative. Some examples are provided in Table S2-1 in Supplementary material 2. As rheumatic fever is usually a historical disease when the patient was a child, the disease was commonly not coded with ICD. 

For certain rare diseases, the accuracy score from the manual evaluation was very low, e.g. 0.0\% for IRIDA syndrome due to ``microcytic anaemia'' wrongly assigned as a synonym or an atom of C0085576 (``Iron-Refractory Iron Deficiency Anemia'' or IRIDA) in the previous UMLS version (2019AA) in the Text-to-UMLS process, 8.2\% and 43.8\% for Retinitis Pigmentosa and Progressive Multifocal Leukoencephalopathy, respectively, due to the ambiguous meanings of their abbreviations (``RP'' and ``PML'') and unseen in WS (with a low corpus-based prevalence below 0.5\%). For Multifocal Atrial Tachycardia, the definition in ORDO is a neonatal disease, while its matched UMLS concept of the same name may also mean an adult disease. We also found difficulty in reaching a consensus in the annotation due to the vague definition of Acute Liver Failure in ORDO\footnote{\url{https://www.ebi.ac.uk/ols/ontologies/ordo/terms?iri=http\%3A\%2F\%2Fwww.orpha.net\%2FORDO\%2FOrphanet_90062}}, for which we derived two distinct interpretations which were then reconciled by a senior clinician\footnote{Our two interpretations of acute liver failure differ most in the factors of drug use, alcohol abuse, virus infection, etc., that could contribute to the rarity of the disease but \underline{not} specified in the definition from ORDO. We finally considered hepatitis virus or drugs as causes of acute liver failure as a rare disease, but removed cases of alcohol abuse.}. This analysis suggests that we should take the definitions into consideration in entity linking and ontology matching. We should also ensure that the definitions used are appropriate for the clinical research question for people using the tools.

Although the accuracy scores were not perfect, for all diseases except IRIDA syndrome, NLP could still enrich the cases identified from ICD-9 after the manual check by the experts. We also find that with ICD codes, it is possible to find cases not identified by NLP as well, as shown in asbestos intoxication, necrotizing enterocolitis, etc., which may be related to the imperfect recall of the NLP model or the rare diseases being not (explicitly) mentioned in the clinical note. In general, the results above on \textit{rare diseases} extend the conclusion of the previous survey in case detection \cite{Ford2016} that NLP with free-texts can greatly enrich the information from ICD codes and the two sources complement each other. We further present the results of NLP with strong supervision in Figure S1-1 in Supplementary material 1, which overall predicted fewer cases and resulted in better accuracy scores, but reflected the same picture as with weak supervision.

\subsection*{Transfer and Re-training with Radiology Reports}
For external validation, we applied the proposed weak supervision pipeline and models to extract rare disease phenotypes from two datasets of radiology reports, US MIMIC-III radiology reports (n=520k) \cite{johnson_mimic-iii_2016} and UK NHS Tayside brain imaging reports (n=156k) \cite{gorinski_named_2019}. For each of the datasets, we selected a subset of clinical notes (1,000 for MIMIC-III and 5000 for Tayside), and obtained the candidate mention-UMLS pairs with SemEHR to be labelled for evaluation. The detailed data statistics are in Table \ref{data-statistics}. Based on the real-world practice of NLP, we consider two ways to apply the pipeline in Figure \ref{pipeline-weak}: (i) model transfer and (ii) in-domain re-training. For model transfer, we directly applied our phenotype confirmation models, $M_{weak}$ (and $M_{strong}$), trained from MIMIC-III discharge summaries to the two new datasets; for in-domain re-training, we created weakly labelled training data from each new dataset and trained a data-specific phenotype confirmation model with Algorithms \ref{WS:data_programming}-\ref{WS:data_rep_train}; we further tuned the parameters $p$ and $l$ in the weak labelling rules during re-training.

Table \ref{external_validation_res} shows the external validation results of the NLP pipeline with model transfer or in-domain re-training. We mainly present the Text-to-UMLS results, consistent with Text-to-ORDO results in Table S1-4 and admission-level results in Table S1-5 in Supplementary material 1. It is observed that directly applying a weak supervision model trained from another type of report (e.g. discharge summaries) could largely improve the precision and $F_1$ score of SemEHR, with a slight drop of recall from nearly 100\% to over 90\%. This transferability of models suggests that there are common linguistic patterns used in all types of clinical notes, even from different sources. The strong supervision model obtained a higher precision, but with a much lower recall (a drop of 20\% to over 30\% compared to SemEHR only) and thus may bear the risk of missing true positive mentions. Results from the in-domain re-training of models were much better than model transfer, as the former could bridge the linguistic gap between discharge summaries and radiology reports even for the same cohort or institution in MIMIC-III. We further tuned the weak labelling parameters to optimise the recall or $F_1$ score. A perfect or no loss of recall (100\% or near 95\%) was achieved on par with SemEHR and the precision was further improved compared to using the original parameters. Although the parameter tuning process was based on the full annotated data, this can be substituted by the inspection of a small number of data at the rule designing stage. Finally, we noticed that simply using rules (SemEHR+rules) with the best tuned parameters was highly effective, achieving better results than most evaluation settings, but still surpassed by the best tuned WS model, especially for the Tayside reports. The results between rules only and weak supervision were consistent with those of the discharge summaries in Table \ref{umls_linking_results}.

\begin{table}[t]
\caption{External Validation Results on Radiology Reports from MIMIC-III and NHS Tayside}
\scriptsize
\center
\label{external_validation_res}
\begin{threeparttable}
\begin{tabular}{llll|lll}
\cline{1-7}
                            & \multicolumn{3}{l}{\begin{tabular}[c]{@{}c@{}}MIMIC-III Radiology\\(n=53+/198)\end{tabular}}         & \multicolumn{3}{l}{\begin{tabular}[c]{@{}c@{}}Tayside Brain Imaging\\(n=79+/283)\end{tabular}}               \\
Text to UMLS                & P            & R              & $F_1$         & P             & R              & $F_1$       \\
\cline{1-7}
SemEHR \cite{Wu2018semehr}  & 26.7        & \textbf{100.0} & 42.2          & 26.9 & \textbf{94.9} & 41.9         \\
+ WS (transfer)             & 54.4         & 92.5           & 68.5          & 56.3 & 91.1           & 69.6     \\
+ SS (transfer)             & \textbf{89.4}& 79.2           & 84.0          & 69.0 & 62.0           & 65.3      \\
\cline{1-7}
+ rules (tuned)                    & 87.5        & 92.5            & 89.9         & 56.8 & 94.9  & 71.1 \\
+ WS (in-domain)            & 72.9         & 96.2           & 82.9          & 48.0 & 92.4           & 63.2     \\
+ WS (+ tuning R)         & 81.5         & \textbf{100.0} & 89.8 & 58.1 & \textbf{94.9}           & 72.1     \\
+ WS (+ tuning $F_1$)         & 89.1         & 92.5 & \textbf{90.7} & \textbf{75.3} & 88.6           &
\textbf{81.4}     \\
\hline\hline
\cline{1-7}
\end{tabular}
\begin{tablenotes}
\item The column statistics (n=$N_+$+/$N$) show number of positive data $N_+$ and all samples $N$ in the dataset. WS, weak supervision; SS, strong supervision. The original parameters for WS were $p=0.005$ and $l=3$. The new parameters for best recall (R) were $p=0.01$ and $l=4$ and for best $F_1$ were $p=0.0005$ and $l=4$, for both datasets. For SemEHR+rules, we present the results of rules, where $p=0.0005$ and $l=4$, with an OR operation. The best scores for the metrics are bolded.
\end{tablenotes}
\end{threeparttable}
\end{table}

\section*{Conclusion, Discussion, and Future Studies}
\label{sec:discussion}

In this study, we proposed an ontology-driven and weakly supervised approach for rare disease phenotyping from clinical notes. Unlike the use of ontologies, weak supervision has not been well established in the clinical NLP domain. Our proposed weak supervised deep learning approach requires no human annotation and extends the paradigm from \cite{wang_clinical_2019} on weak supervision for clinical texts, by introducing ontologies, named entity linking tools, and contextual representations. We designed two simple but effective rules (mention character length and corpus-based ``prevalence'') to create weakly labelled data regarding ambiguous abbreviations and rare entities. The trained phenotype confirmation model effectively filtered out the false positives in the data with no (or a minimum) side effect on the true positives. 

Traditional clinical NLP relies heavily on strong supervision with manually labelled data. However, with recent data-demanding methods like deep learning, it is time to consider to automatically create labelled data to train models, with the support of rules and resources like ontologies and NER+L tools. Our work on rare diseases provides empirical evidence for the task by applying a weakly supervised NLP pipeline on three clinical note datasets (one for discharge summaries and two for radiology reports) in two institutions in the US and the UK. The improvements on the precision were highly significant (by over 30\% to 50\% absolute score for Text-to-UMLS linking), with almost no loss of recall compared to the existing NER+L tool, SemEHR. Our study also demonstrates that NLP can complement traditional ICD-based approaches to better estimate rare diseases in clinical notes (see Figure \ref{patient_icd_vs_nlp}).

While our rule-based weak supervision does not require annotated data, it can bring bias or noise as no simple rule can perfectly predict the labels for a complex task. This bias, although not affecting most predictions for the testing data, was manifested in the slight drop of recall in Text-to-UMLS linking (Table \ref{umls_linking_results}). This loss of recall may be minimised through tuning the parameters in the weak labelling rule (e.g. relaxing the ``prevalence'' or mention length threshold, shown in Table \ref{external_validation_res}), but needs a small set of annotated data or some manual inspection of the predictions. The mention character length rule may also be enhanced with accurate abbreviation expansion and disambiguation to retain abbreviations that are rare diseases. Besides, recent studies in the general NLP domain have begun tackling the bias of rules (with a rule-level attention mechanism \cite{karamanolakis2021}) or noise of weakly labelled data (with the estimation of data-level confidence \cite{jiang2021}). Also, we used a heuristic-based logic operation (as XNOR) to aggregate the two rules; future studies can explore more advanced aggregation methods (e.g., learning a label model \cite{fries2021ontology,ratner2020snorkel}).

As suggested in our results and other studies \cite{fries2021ontology,ratner2020snorkel}, the current performance of the best weakly supervised methods is still below strong supervision. But the gap between the weak and strong supervision is small (within 5\% $F_1$ score) and there is no difference in terms of recall. This shows that the expensive and time-consuming annotations for text phenotyping may be greatly reduced, substituted by an alerting system or manual screening based on the predictions of a weakly supervised NLP system. With a small number of annotated data for parameter tuning, both the precision and recall of our weak NLP model were further improved (see Table \ref{external_validation_res}). This may suggest a future study to better use a small sample of annotated data with the weakly annotated data for semi-supervised learning to improve the performance.

There are still, however, some false positive mentions detected by the proposed NLP pipeline, as shown in our analyses of the prediction errors and the identified cohorts (in Figures \ref{error_pie_chart}-\ref{patient_icd_vs_nlp}). Disambiguating entity types (especially for abbreviations) still remains a challenge for text phenotying. This suggests to potentially integrate word sense disambiguation to enhance the weak supervision approach, e.g., through more reliable weak data creation. Also, errors in identifying hypothetical and negation (``Hypo/neg'') mentions suggest to separately model ``Hypo/neg'' in the classification, which can be learned with mentions beyond the scope of rare diseases. Furthermore, the complexities of linguistic patterns of a (rare) disease may still require better representations beyond the current context window and may need to be enhanced with ontology concepts. Our evaluation of the NLP-identified cases suggests modelling the semantics of the lexical definitions in ontologies (e.g. ORDO) to improve entity linking and ontology matching.

Also, we note that our work is highly dependent on existing ontologies and their available matchings to each other. We leveraged and validated the matching among ORDO, UMLS, ICD-10, and ICD-9. The current matchings are generally correct, but not perfect (e.g. 88.4\% accuracy of matching between UMLS and ORDO). A more accurate matching among ontologies, potentially corrected with machine learning \cite{kolyvakis2018}, will improve the performance of our pipeline. It is also possible to directly match texts to ORDO, which can include rare diseases not contained in UMLS, but this does not leverage the synonyms in UMLS that represent the name variation of rare disease entities. Also, our approach cannot identify emerging rare disease entities, not contained in the ontologies and not thus easily captured by SemEHR, which is the next, challenging direction for our study.

While we only enhanced SemEHR with the weakly supervised phenotype confirmation model, the approach can be adapted to improve other NER+L tools and models to support more accurate rare disease cohort selection and coding. Recently, more packages and environments (e.g. Snorkel \cite{ratner2020snorkel}, skweak \cite{lison2021}) have been created to apply weak supervision in general domain NLP practice. Thus, a promising future study is to adapt the current weak supervision infrastructures or the ideas behind them to the clinical NLP domain and establish best practices in the field; a recent work adapting Snorkel \cite{ratner2020snorkel} is Trove \cite{fries2021ontology}, which has not yet been applied to the domain of rare diseases, that involves additional ontologies and their mappings.

Our work mainly focused on identifying rare disease concepts in the clinical notes, while other physical, behavioural, and physiological characteristics need to be identified so as to establish a clinical diagnosis of a rare disease. We also mainly focused on rare diseases as a whole and the approach can be applied to identify specific rare diseases. Future work needs to extract a wider set of information to enhance rare disease phenotyping, and to facilitate the development of risk prediction tools for rare diseases to support decision making during the COVID-19 pandemic and beyond \cite{zhang2021,zhang2022}.

\begin{backmatter}
\section*{Abbreviations}
BERT: Bidirectional Encoder Representations from Transformers; UMLS: Unified Medical Language System; ORDO: Orphanet Rare Disease Ontology; MIMIC-III: Medical Information Mart for Intensive Care; NHS: National Health Services; NLP: Natural Language Processing; NER+L: Named Entity Recognition and Linking; ICD: International Classification of Diseases; WS: Weak Supervision; SS: Strong Supervision.
\end{backmatter}

\section*{Declarations}
\begin{backmatter}
\section*{Acknowledgement}
This work is a substantial extension of our previous work in \cite{dong2021rare}, which provided the first versions of Figures \ref{pipeline-main} and \ref{pipeline-weak} (re-created and revised in this paper) and some preliminary results on MIMIC-III discharge summaries. We would like to thank Emma Whitfield for the important support on data annotations of discharge summaries during the previous study \cite{dong2021rare} and feedback on the writing of this work. This work has made use of the resources provided by the Edinburgh Compute and Data Facility (ECDF).

\section*{Funding}
This work is supported by Health Data Research UK National Phenomics and Text Analytics Implementation Projects, Wellcome Institutional Translation Partnership Awards (PIII009, PIII029, PIII032, PIII054), Medical Research Council and Health Data Research UK (MR/S004149/1). HZ and AC are supported by the Advanced Care Research Centre (ACRC). HD and JC are also supported by EPSRC project ConCur on Knowledge Graph Construction and Curation (EP/V050869/1). The funding bodies are independent of the design of the study and collection, analysis, and interpretation of data and in writing the manuscript.

\section*{Availability of data and materials}
The MIMIC III datasets are available at \url{https://mimic.physionet.org/} upon request after the ethical training. NHS Tayside data are not publicly available due to the privacy of patients and please refer to \url{https://www.dundee.ac.uk/hic} regarding further interest in the dataset. The rare disease mention annotations of MIMIC-III discharge summaries and radiology reports, along with the implementation of the approach, are available at \url{https://github.com/acadTags/Rare-disease-identification}.

\section*{Ethics approval and consent to participate}
We were granted access to MIMIC-III through PhysioNet after completing the ethical training in human research subject protections and HIPAA regulations, through the Collaborative Institutional Training Initiative program (\url{https://physionet.org/about/citi-course/}). We have also received NHS Tayside Caldicott Guardian approval (CSAppMW1758) to use the anonymised brain imaging reports for this work. All our methods were carried out in accordance with relevant guidelines and regulations. The approval of both MIMIC-III and NHS Tayside datasets allows us to carry out Natural Language Processing experiments on the reports. All reports have been de-identified and we do not identify any individual patients in the methods and experiments, thus the research is exempt from requiring informed consent from the patients according to the NHS Tayside Caldicott Guardian approval. 

\section*{Competing interests}
The authors declare that they have no competing interests.

\section*{Consent for publication}
Not applicable.

\section*{Authors' contributions}
HD, HW, VSP, HZ, and MW conceptualised the research. HD, HW, VSP, MW, and HZ designed the method and experiments. HD, MW, and VSP implemented the approach. MW, HZ, ED, AC, and other researchers annotated the datasets or screened the detected rare disease cases. ED and WW provided clinical suggestions on screening the detected rare disease cases. WW and BA applied for the ethical approval for data access to NHS Tayside brain imaging reports. BA established the secure data server for experimentation. JC provided feedback on ontology-based methods and revisions. HD drafted the paper. All authors read and revised the draft and approved the final manuscript.


\bibliographystyle{bmc-mathphys}
\bibliography{rare_disease_coding_res}
\end{backmatter}
\end{document}


\begin{frontmatter}
		
		\title{Supplementary materials 1 - Further Parameter Settings and Results and 2 - Examples of Rare Disease Identification and Ontology Matching}
	\end{frontmatter}
	
	\section*{Weak Rule Parameter Tuning}
	\label{appendix:param-tuning}
	The results of parameter tuning for weak labelling rules regarding $F_1$, recall, and precision scores, are displayed in Table \ref{umls_linking_f1_p_r_tuning}. We tuned through a grid search the possible values of $p \in \{1\mathrm{e}{-4},5\mathrm{e}{-4},1\mathrm{e}{-3},5\mathrm{e}{-3},1\mathrm{e}{-2},5\mathrm{e}{-2},1\mathrm{e}{-1}\}$ and $l \in \{2,3,4\}$, and selected the model based on recall and $F_1$ scores in Text-to-UMLS linking. For MIMIC-III discharge summaries, the results were based on the 400 validation set of manually annotated mention-UMLS pairs. The parameter $p$ controls the corpus-based ``prevalence'' of the disease concept, which is related to epidemiological information, e.g., the actual prevalence of a rare disease in the cohort. A higher $p$ resulted in more disease concepts selected, thus higher recall but generally less precision. The parameter $l$ controls the mention length, a key threshold to filter out abbreviations, which are usually ambiguous in their meanings. A higher $l$ thus generally resulted in a higher precision but lower recall. We observed that a corpus-based ``prevalence'' threshold of 0.005 and a mention character length threshold of 3 resulted in the best $F_1$ score for the dataset. We thus recommend to set $p \in \{0.005,0.01\}$ and $l \in \{3,4\}$ and used $p$ as 0.005 and $l$ as 3 for MIMIC-III discharge summaries. 
	
	Also, we found that the results were not sensitive to window size, i.e. the input tokens before and after the mention. Thus, all models used the default window size of 5.
	
	\begin{table*}[ht]
		\caption{$F_1$, Precision (P), and Recall (R) scores with respect to the weak rule parameters $p$ and $l$ in Text-to-UMLS linking for MIMIC-III discharge summaries (with the highest $F_1$ score in bold)}
		\center
		\scriptsize
		\label{umls_linking_f1_p_r_tuning}
		\begin{tabular}{llll|lll|lll}
			\cline{1-10}
			\multirow{2}{*}{} & \multicolumn{3}{c|}{$l$ = 2}                                              & \multicolumn{3}{c|}{$l$ = 3}                                              & \multicolumn{3}{c}{$l$ = 4}                                              \\
			& \multicolumn{1}{c}{$F_1$} & \multicolumn{1}{c}{P} & \multicolumn{1}{c|}{R} & \multicolumn{1}{c}{$F_1$} & \multicolumn{1}{c}{P} & \multicolumn{1}{c|}{R} & \multicolumn{1}{c}{$F_1$} & \multicolumn{1}{c}{P} & \multicolumn{1}{c}{R} \\
			\cline{1-10}
			$p$ = 0.0001        & 73.7\%                 & 64.1\%                & 86.6\%                & 59.3\%                 & 86.5\%                & 45.1\%                & 64.3\%                 & 87.8\%                & 50.7\%                \\
			$p$ = 0.0005        & 73.7\%                 & 59.6\%                & 96.5\%                & 82.6\%                 & 91.5\%                & 75.4\%                & 79.8\%                 & 91.0\%                & 71.1\%                \\
			$p$ = 0.001         & 72.8\%                 & 57.3\%                & 100.0\%               & 80.8\%                 & 91.2\%                & 72.5\%                & 79.8\%                 & 91.0\%                & 71.1\%                \\
			$p$ = 0.005         & 71.5\%                 & 55.7\%                & 100.0\%               & \textbf{89.8\%}                 & 90.1\%                & 89.4\%                & 87.5\%                 & 91.5\%                & 83.8\%                \\
			$p$ = 0.01          & 71.0\%                 & 55.0\%                & 100.0\%               & 89.7\%                 & 87.3\%                & 92.3\%                & 88.5\%                 & 90.4\%                & 86.6\%                \\
			$p$ = 0.05          & 63.7\%                 & 46.7\%                & 100.0\%               & 64.0\%                 & 47.0\%                & 100.0\%               & 64.3\%                 & 47.3\%                & 100.0\%               \\
			$p$ = 0.1           & 60.0\%                 & 42.9\%                & 100.0\%               & 60.0\%                 & 42.9\%                & 100.0\%               & 60.0\%                 & 42.9\%                & 100.0\%              \\
			\cline{1-10}
		\end{tabular}
	\end{table*}
	
	\section*{Embedding and Fine-tuning Settings}\label{appendix:emb-hyper-param}
	We controlled the same window size (as 5) in baselines with word2vec embeddings and BERT model fine-tuning.
	
	For the word2vec embeddings pre-trained on MIMIC-III discharge summaries, we used Gensim library\footnote{\url{https://radimrehurek.com/gensim/auto_examples/tutorials/run_word2vec.html}} with Continuous Bag of Words algorithm, without filtering vocabularies by frequency (min\_count=0). We experimented with the dimensions of 100, 300, and 768 (see Table 4 in the paper).
	
	For fine-tuning BERT models, we used the average pooling of the mention's sub-tokens representations in the second-last layer (same as the Contextual Mention Representation, with fine-tuning instead of static embedding), followed by a linear layer with softmax activation and cross-entropy loss. The learning rate, warmup steps, and weight decay were 5e-05, 500, and 0.01, resp., set up using Huggingface Trainer\footnote{\url{https://huggingface.co/docs/transformers/main_classes/trainer}}, and trained with 3 epochs. We fine-tuned the BlueBERT-based model (see Table 4 in the paper).
	
	\section*{Results on Different Encoding Strategies}
	\label{appendix:encoding}
	The first encoding strategy is \emph{mention masking}, whether or not to mask the mention in the full context window. The intuition behind this is to explore the potential of a language model to confirm a phenotype solely based on the surrounding context but not the mention itself.
	
	The second encoding strategy is \textit{using document structure names} (or template section names) to enhance local context. If the document structure name $s$ is available in the dataset, we add $s$ before the context window $t$ with a separation token [SEP] in between.
	
	Results on the different encoding strategies for Text-to-UMLS linking in MIMIC-III discharge summaries are displayed in Table \ref{umls_linking_encoding_results}. Non-masked encoding achieved better results than masked encoding. Using document structure names further boosted recall scores on the validation and the test set. We used non-masked encoding (with document structure names for MIMIC-III discharge summaries only) for data representation.
	
	\begin{table}[ht]
		\caption{Comparison among encoding strategies for weakly supervised Text-to-UMLS linking on MIMIC-III discharge summaries}
		\scriptsize
		\center
		\label{umls_linking_encoding_results}
		\begin{threeparttable}
			\begin{tabular}{llll|lll}
				\cline{1-7}
				& \multicolumn{3}{c}{validation set (n=142+/400)}         & \multicolumn{3}{c}{test set (n=187+/673) } \\
				Text to UMLS                & P          & R          & $F_1$ & P          & R          & $F_1$     \\
				\cline{1-7}
				non-M     & 89.9        & 87.3        & 88.6  & \textbf{81.3}       & 90.9       & \textbf{85.9}     \\
				non-M+DS  & \textbf{90.1}        & \textbf{89.4}        & \textbf{89.8} & 80.4       & \textbf{92.0}       & 85.8       \\
				M         & 86.5        & 63.4        & 73.2 & 78.6       & 61.0       & 68.7      \\
				M+DS      & 86.4        & 62.7        & 72.7 & 78.0       & 62.6       & 69.4           \\
				\cline{1-7}
			\end{tabular}
			\begin{tablenotes}
				\item M denotes mention \underline{m}asking and non-M denotes no mention masking applied. DS denotes using \underline{d}ocument \underline{s}tructure names. The non-M+DS model was trained on the full set of weakly labelled data, without tuning the optimal number of data, thus slightly below results in Table 2. BlueBERT-base (PubMed+MIMIC-III) was used to encode the text sequences.
			\end{tablenotes}
		\end{threeparttable}
	\end{table}
	
	\section*{NLP with Strong Supervision vs. ICD for Admission-level Rare Disease Identification}
	\label{appendix:ss-results}
\begin{figure*}[t]
		\centering
		\includegraphics[width=0.99\textwidth]{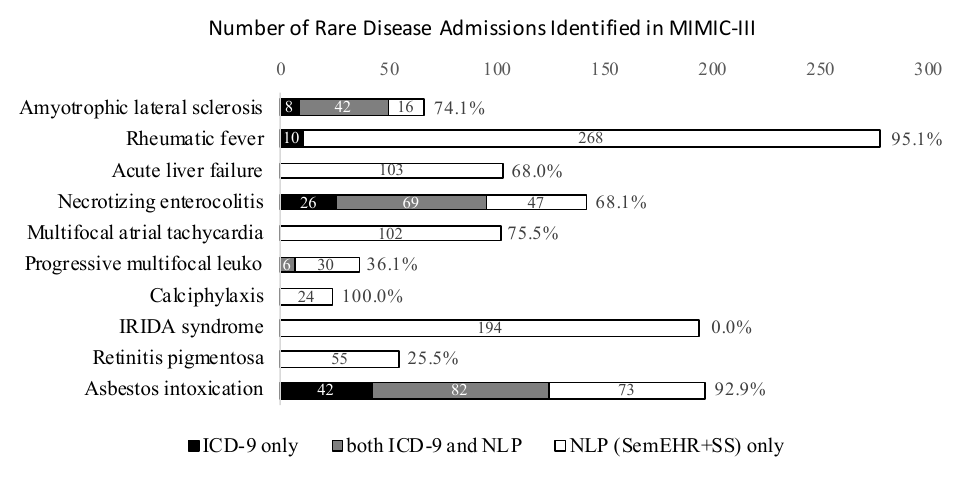}
		\caption{Number of rare disease patient stays from MIMIC-III (n=59,652): ICD (code-based) vs. NLP (text-based, with \textit{strong} supervision). The 10 rare diseases are the same as those presented for weak supervision in Figure 4 in the paper. Admissions are split into those \textit{only} identified through links from ICD-9 codes (in black), those \textit{only} identified from free texts with strong supervision (NLP, in white), and the intersection of cases from both ICD-9 and NLP (in grey). The percentage after each horizontal bar shows the accuracy of NLP based on the manual assessment of the identified cases.}\label{patient_lvl_SS}
	\end{figure*}

	Figure \ref{patient_lvl_SS} shows the results of the NLP pipeline with strong supervision compared to ICD codes for admission-level rare disease phenotyping. The results were generally consistent with the weak supervision approach (in Figure 4 in the paper) that NLP-based results greatly complement the code-based rare disease cohort. Generally, a higher accuracy with a less number of admissions was predicted by strong supervision compared to weak supervision (e.g. the accuracy was 25.5\% or 14/55 predicted by ``Retinitis Pigmentosa'' for strong supervision, compared to 8.2\% or 15/183 predicted by weak supervision).
	
	\section*{Overall Admission-level and Mention-level Results}
	\label{appendix:overall-results}
	Table \ref{admission_rd_id_results} shows the admission-level rare disease phenotyping results for MIMIC-III discharge summaries.
	
	\begin{table}[t]
		\caption{\underline{Micro-level} results of admission-level rare disease identification for MIMIC-III discharge summaries}
		\scriptsize
		\center
		\label{admission_rd_id_results}
		\begin{threeparttable}
			\begin{tabular}{llll|lll}
				\cline{1-7}
				& \multicolumn{3}{l}{\begin{tabular}[c]{@{}c@{}}validation\\(n=30+/117$*$55)\end{tabular}}         & \multicolumn{3}{l}{\begin{tabular}[c]{@{}c@{}}test\\(n=42+/192$*$82)\end{tabular}}               \\
				Admission to ORDO          & P             & R              & $F_1$         & P             & R              & $F_1$       \\
				\cline{1-7}
				SemEHR  & 15.4          & \textbf{93.3}  & 26.4          & 12.7          & \textbf{95.2}  & 22.3         \\
				+ rules                     & 39.2          & 66.7           & 49.4          & 38.9          & 88.1           & 54.0               \\
				+ WS (rules+BERT)           & \textbf{57.1} & 66.7           & \textbf{61.5} & \textbf{49.3} & 78.6           & \textbf{60.6}     \\
				+ SS (anns+BERT)            & -             & -              & -             & \textbf{61.1} & 78.6           & \textbf{68.7}      \\
				\hline\hline
				ICD                         & 56.2          & 30.0           & 39.1          & 27.3          & 21.4           & 24.0      \\
				ICD $\cup$ SemEHR+WS        & 50.0          & 70.0           & 58.3          & 40.4          & 85.7           & 55.0      \\
				ICD $\cup$ SemEHR+SS        & -             & -              & -             & 45.9          & 81.0           & 58.6      \\
				\cline{1-7}
			\end{tabular}
			\begin{tablenotes}
				\item The micro-level metric counts each admission and an associated ORDO concept (or an admission-ORDO pair) as a single instance. The column statistics (n=$N_+$+/$N_d*N_l$) show the number of positive data $N_+$, admissions (or discharge summaries) $N_d$, and unique candidate rare diseases (or ORDO concepts) $N_l$ in the dataset. WS, weak supervision; SS, strong supervision; anns, annotations. BlueBERT-base (PubMed+MIMIC-III) was used as the BERT model. ICD denotes the approach to matching ICD-9 codes to ORDO concepts. The union sign ($\cup$) denotes merging and de-duplicating the cases identified from the two methods. Precision (P) and $F_1$ for ICD-based methods may be lower than actual values, as all candidate mentions were from SemEHR.
			\end{tablenotes}
		\end{threeparttable}
	\end{table}
	
	Table \ref{external_validation_res_T2O} and \ref{external_validation_res_A2O} show the overall mention-level (Text-to-ORDO) and admission-level results of two radiology report datasets in the US (MIMIC-III) and the UK (NHS Tayside). For Tayside data, the recall was lower as we manually identified new rare disease mentions that were not included in the candidate mentions from SemEHR. Weak supervision (WS) achieved better recall than transferring the SS model in results from both Tables. The code-based approach (ICD) also did not show an advantage in identifying more rare disease admissions (see recall, R), and overall performance (see $F_1$), comparing ICD or ``ICD $\cup$ SemEHR+WS'' with the (best) SemEHR+WS setting in Table  \ref{external_validation_res_A2O} and Table \ref{admission_rd_id_results}, but the results may be biased towards methods adapting SemEHR as it was used as a starting source to create candidate mentions for the manual annotation.
	
	\begin{table}[t]
		\caption{Results on rare disease identification (Text-to-ORDO) for MIMIC-III and Tayside radiology reports}
		\center
		\scriptsize
		\label{external_validation_res_T2O}
		\begin{threeparttable}
			\begin{tabular}{llll|lll}
				\cline{1-7}
				& \multicolumn{3}{l}{\begin{tabular}[c]{@{}c@{}}MIMIC-III Radiology\\(n=46+/198)\end{tabular}}         & \multicolumn{3}{l}{\begin{tabular}[c]{@{}c@{}}Tayside Brain Imaging\\(n=42+/283)\end{tabular}}               \\
				Text to ORDO                & P            & R              & $F_1$         & P             & R              & $F_1$       \\
				\cline{1-7}
				SemEHR  & 22.9        & \textbf{93.5} & 36.8            & 13.1 & \textbf{78.6} & 22.4         \\
				+ WS (transfer)             & 48.8         & 84.8           & 61.9          & 31.4 & 76.2           & 44.4     \\
				+ SS (transfer)             & 86.5         & 69.6           & 77.1          & \textbf{53.2} & 59.5           & 56.2      \\
				\cline{1-7}
				+ rules (tuned)             & 84.8        & 84.8          & 84.8            & 31.4         & 76.2           & 44.4     \\
				+ WS (in-domain)            & 68.3         & 89.1           & 77.4          & 26.4 & \textbf{78.6}           & 39.5     \\
				+ WS (+ tuning R)           & 78.2         & \textbf{93.5}  & 85.1          & 32.4 & \textbf{78.6}           & 45.8     \\
				+ WS (+ tuning $F_1$)       & \textbf{86.7}         & 84.8  & \textbf{85.7}          & 46.3 & 73.8                    & \textbf{56.9}    \\ \hline\hline
				\cline{1-7}
			\end{tabular}
			\begin{tablenotes}
				\item The column statistics (n=$N_+$+/$N$) shows the number of positive data $N_+$ and the overall number of samples $N$ in the dataset. WS, weak supervision; SS, strong supervision. The original parameters for WS were $p=0.005$ and $l=3$. The new parameters for best recall (R) were $p=0.01$ and $l=4$ and for best $F_1$ were $p=0.0005$ and $l=4$, for both datasets. For SemEHR+rules, rules were aggregated with an OR operation and $p=0.0005$ and $l=4$.
			\end{tablenotes}
		\end{threeparttable}
	\end{table}
	
	\begin{table}[t]
		\caption{\underline{Micro-level} results of admission-level rare disease identification for MIMIC-III and Tayside Radiology Reports}
		\center
		\scriptsize
		\label{external_validation_res_A2O}
		\begin{threeparttable}
			\begin{tabular}{llll|lll}
				\cline{1-7}
				& \multicolumn{3}{l}{\begin{tabular}[c]{@{}c@{}}MIMIC-III Radiology\\(n=29+/145$*$43)\end{tabular}}         & \multicolumn{3}{l}{\begin{tabular}[c]{@{}c@{}}Tayside Brain Imaging\\(n=41+/273$*$65)\end{tabular}}               \\
				Admission to ORDOs          & P             & R              & $F_1$         & P             & R              & $F_1$       \\
				\cline{1-7}
				SemEHR                      & 19.4        & \textbf{93.1} & 32.1            & 12.8 & \textbf{78.0} & 22.0         \\
				+ WS (transfer)             & 38.7         & 82.8           & 52.7          & 30.7 & 75.6           & 43.7     \\
				+ SS (transfer)             & \textbf{83.3}         & 69.0           & 75.5          & \textbf{53.2} & 61.0           & 56.8      \\
				\cline{1-7}
				+ rules (tuned)     & 80.0        & 82.8          & 81.4            & 30.7         & 75.6           & 43.7     \\
				+ WS (in-domain)            & 59.5         & 86.2           & 70.4         & 25.8 & \textbf{78.0}           & 38.8     \\
				+ WS (+ tuning R)           & 71.1         & \textbf{93.1}  & 80.6          & 31.7 & \textbf{78.0}           & 45.1  \\
				+ WS (+ tuning $F_1$)       & 82.8         & 82.8  & \textbf{82.8}          & 46.3 & 75.6           & \textbf{57.4}     \\
				\hline\hline
				ICD                         & 46.4          & 44.8           & 45.6          & -             & -              & -      \\
				ICD $\cup$ SemEHR+WS        & 51.9          & \textbf{93.1}           & 66.7          & -             & -              & -      \\
				\cline{1-7}
			\end{tabular}
			\begin{tablenotes}
				\item The micro-level metric counts each admission and an associated ORDO concept (or an admission-ORDO pair) as a single instance. The column statistics (n=$N_+$+/$N_d*N_l$) show the number of positive data $N_+$, the number of admissions (or discharge summaries) $N_d$, and the number of candidate rare diseases (or ORDO concepts) $N_l$ in the dataset. WS, weak supervision; SS, strong supervision. The original parameters for WS were $p=0.005$ and $l=3$. The new parameters for best recall (R) were $p=0.01$ and $l=4$ and for best $F_1$ were $p=0.0005$ and $l=4$, for both datasets. For SemEHR+rules, rules were aggregated with an OR operation and $p=0.0005$ and $l=4$. The union sign ($\cup$) denotes merging and de-duplicating the cases identified from the two methods. For ICD $\cup$ SemEHR+WS, the WS model was ``in-domain + tuning R'', the one re-trained with in-domain data and optimised recall. Precision (P) and $F_1$ for ICD-based methods may be lower than actual values, as all candidate mentions were from SemEHR.
			\end{tablenotes}
		\end{threeparttable}
	\end{table}
	
	\renewcommand{\thetable}{S2-\arabic{table}}
	\renewcommand{\thefigure}{S2-\arabic{figure}}
	
	\section*{Examples of Rare Disease Text Phenotyping}
	\label{appendix:examples}
	
	Table \ref{error_analysis_example} (on page 5) shows some selected prediction errors and a few correct predictions. The first four examples are the false positives selected in the evaluation data for the weak supervision model due to semantic type errors, hypothetical contexts, or other issues. The last five examples are those selected from the identified rare disease cohort for Retinitis Pigmentosa and Rheumatic Fever. Synonyms in UMLS could help identify some name variations, e.g. ``tracheobronchomalacia'' for Williams-Campbell syndrome, and ``acute rheumatic fever'' for Rheumatic fever, but also introduces false positives especially regarding abbreviations, e.g. ``EMA'' and ``RP''. The complex context in the clinical notes, including the relative's diseases or hypothetical mentions, although only representing a small part of cases, were still challenging for the NLP pipeline (SemEHR+WS), as these were not explicitly considered in the weakly supervised training process. We also note that there were errors in parsing the document structure name through regular expressions in SemEHR, which might affect the predictions.
	
	\section*{Ontology Matching from ORDO to ICD-9}
	
	Table \ref{onto-matching-ordo2icd} (on page 6) shows 10 examples of rare disease concepts and their ontology matching from ORDO to UMLS, ICD-10, and ICD-9. The rare diseases are the same as those presented in Figure 4 in the paper and Figure S1-1 in Supplementary material 1.
	
	\begin{table*}[ht]
		\caption{Examples of wrong and correct rare diseases identified by SemEHR with the weak supervised phenotype confirmation model from MIMIC-III discharge summaries}
		\scriptsize
		\center
		\label{error_analysis_example}
		\begin{threeparttable}
			\begin{tabular}{lp{2.3cm}p{2.9cm}p{1cm}p{1.8cm}p{0.3cm}p{0.3cm}p{2cm}}
				\cline{1-8}
				ROW\_ID & Document Structure         & Text (with \textbf{mention} in bold) & UMLS     & ORDO   & Pred & Label & Potential Reason                  \\
				\cline{1-8}
				26825   & pertinent\_results (should be pathology)         & Pathology: ...Immunostains for cytokeratin AE1/3 and CAM 5.2, CD-68, CD-79a, CD-138, S-100, LCA absorbed CEA, \textbf{EMA}, CD34, CD31, TTF-1, actin, desmin, MNF-116, calcitonin, and thyroglobulin are negative... & C0268596 & 26791  Multiple acyl-CoA dehydrogenase deficiency & \textcolor{red}{T} & F & negation with a long context, ambiguous mention (EMA as epithelial membrane antigen), and semantic type error (negative test)           \\
				\cline{1-8}
				869     & Hospital\_course           & Brief Hospital Course: \#\# Dyspnea - ...DFA for flu was negative; urinary \textbf{legionella} antigen was also negative... & C0023241 & 549  Legionellosis                                  & \textcolor{red}{T}       & F & semantic type error with negation (negative test) \\
				\cline{1-8}
				8960    & History\_of\_Past\_Illness & Past Medical History: 1. Diagnosed in his early years with bilateral uveitis, clinically had bilateral uveitis significant with loss of vision and \textbf{sarcoid} floaters in both eyes… & C0036202 & 797  Sarcoidosis                                    & \textcolor{red}{T}       & F & not enough information (sarcoid floater not necessary means sarcoid)           \\
				\cline{1-8}
				46361   & pertinent\_results (should be impression)         & IMPRESSION: ...Of note prior chest CT scans have findings suggesting a propensity to \textbf{tracheobronchomalacia}, as well as moderately severe emphysema.... & C0340231 & 411501    Williams-Campbell syndrome                & \textcolor{red}{T}       & F & hypothetical context              \\
				\cline{1-8}
				48161   & Admission\_Medications     & Medications on Admission: ...Vitamin A palmitate 100,000 units 1.5 tablets dialy for \textbf{retinitis pigmentosa}, acetaminophen, tums, Mylanta, OTC   Prilosec prn & C0035334 & 791 Retinitis Pigmentosa                            & T       & T  & correct                                  \\
				\cline{1-8}
				26351   & Hospital\_course           & ...Her ASA continued to be held due to the \textbf{RP} bleed but was restarted after 48 hrs of stable Hct…  & C0035334 & 791 Retinitis Pigmentosa                            & \textcolor{red}{T}       & F & ambiguous abbreviation (Retroperitoneal bleeding)\\
				\cline{1-8}
				12659   & History\_of\_Past\_Illness & Past Medical History: PMHx: ... 7. h/o of \textbf{rheumatic fever} with Sydenham's   chorea… & C0035436 & 3099 Rheumatic fever                                & T       & T  & correct        \\
				\cline{1-8}
				20984   & Hospital\_course           & The patient never reported any pharyngitis, but given his complaints of   diffuse arthralgias, myalgias, migrating neuropathic pain, there was some   concern of \textbf{rheumatic fever}, as the patient had 2 ASO screens performed which   were both negative. & C0035436 & 3099 Rheumatic fever                                & \textcolor{red}{T}       & F & hypothetical context              \\
				\cline{1-8}
				11568   & basic (should be family history)                     & FAMILY HISTORY:  ...2) His mother has an enlarged heart which may be secondary to a   history of \textbf{acute rheumatic fever}... & C0035436 & 3099 Rheumatic fever                                & \textcolor{red}{T}       & F & a relative's disease             \\
				\cline{1-8}
			\end{tabular}
			\begin{tablenotes}
				\item Prediction errors are coloured with \textcolor{red}{red} in ``Pred'' (third-last) column. For columns ``Pred'' and ``Label'', ``T'' means that the prediction or gold is \emph{True} and ``F'' indicates \emph{False}. The wrongly parsed document structure names in the second column are marked with corrected ones in the form of ``(should be XXX)''.
			\end{tablenotes}
		\end{threeparttable}
	\end{table*}
	
	\begin{table*}[ht]
		\caption{Ontology concept matching among ORDO, UMLS, ICD-10, and ICD-9 based on publicly available sources}
		\scriptsize
		\center
		\label{onto-matching-ordo2icd}
		\begin{threeparttable}
			\begin{tabular}{lp{2.2cm}lp{0.8cm}|p{1.1cm}p{2.2cm}|p{1.1cm}p{2.2cm}}
				\cline{1-8}
				ORDO             & ORDO Preferred Label                         & UMLS     & ICD-10                              & ICD-9-NZ (from ICD-10)                                 & Preferred Label                                                                                                                                                                                                                      & ICD-9-BP (from UMLS)   & Preferred Label                              \\
				\cline{1-8}
				803    & Amyotrophic lateral sclerosis              & C0002736 & $<$G122                                & -                                   & -                                                                                                                                                                                                         & 335.20      & Amyotrophic lateral sclerosis              \\
				3099   & Rheumatic fever                              & C0035436 & $>$I011, $>$I00, $>$I010, $>$I012, $>$I018, $>$I019 & 3911, 390,  3910,   3912, 3918, 3919 & Acute rheumatic endocarditis; Rheumatic fever without mention of heart involvement; Acute rheumatic pericarditis; Acute rheumatic myocarditis; Other acute rheumatic heart disease; Acute rheumatic heart disease, unspecified & 390-392.99 & ACUTE RHEUMATIC FEVER                        \\
				90062  & Acute liver failure                          & C0162557 & $<$K720                                & -                                      & -                                                                                                                                                                                                 & -           & -                                             \\
				391673 & Necrotizing enterocolitis                  & C0520459 & $=$P77                                 & 7775                                     & Necrotizing enterocolitis in newborn                                                                                                                                                                                                                                     & -           & -                                             \\
				3282   & Multifocal atrial tachycardia                & C0221158 & $<$I471                                & -                                     & -                                                                                                                                                                                              & -           &  -                                            \\
				217260 & Progressive multifocal leukoencephalopathy & C0023524 & $=$A812                                & 0463                                      & Progressive multifocal leukoencephalopathy                                                                                                                                                                                         & 46.3       & Progressive multifocal leukoencephalopathy \\
				280062 & Calciphylaxis                                & C0006666 & $<$E835                                & -                                     & -                                                                                                                                                                                                                                     &  -          & -                                             \\
				791    & Retinitis pigmentosa                         & C0035334 & $<$H355                                & -                                    & -                                                                                                                                                                                          &  -          &  -                                            \\
				209981 & IRIDA syndrome                               & C0085576 & $<$D508                                & -                                     & -                                                                                                                                                                                            & -           &  -                                            \\
				2302   & Asbestos intoxication                        & C0003949 & $<$J61                                 & -                                      & -                                                                                                                                                                                                                           & 501        & Asbestosis                                   \\
				\cline{1-8}
			\end{tabular}
			\begin{tablenotes}
				\item $=$, $>$, and $<$ in ORDO-to-ICD-10 mappings (all from ORDO) indicate exact, broader-to-narrower, and narrower-to-broader matching, respectively. Narrower-to-broader matching ($<$) from ORDO to ICD-10 was not used for phenotyping, as it may result in common or non-rare diseases' ICD codes. All ORDO-to-UMLS mappings  (all from ORDO) indicate exact matching ($=$). ``ICD-9-NZ'' denotes the set of ICD-9 codes linked from ICD-10 codes using the matching from the Ministry of Health, New Zealand, \url{https://www.health.govt.nz/nz-health-statistics/data-references/mapping-tools/mapping-between-icd-10-and-icd-9}. ``ICD-9-BP'' refers to the set of ICD-9 codes linked from UMLS based on the ICD-9-CM ontology (version 2020AB) in BioPortal, \url{https://bioportal.bioontology.org/ontologies/ICD9CM}.
			\end{tablenotes}
		\end{threeparttable}
	\end{table*}